\documentclass{article}
\usepackage[T1]{fontenc} % add special characters (e.g., umlaute)
\usepackage[utf8]{inputenc} % set utf-8 as default input encoding
\usepackage{ismir,amsmath,cite,url}
\usepackage{graphicx}
\usepackage{dblfloatfix}
\usepackage{color}
\graphicspath{ {./figs/} }
\usepackage[export]{adjustbox}
\usepackage{lineno}
\usepackage[normalem]{ulem}
\usepackage{subcaption}
\usepackage{algorithmic}
\usepackage{algorithm}
\usepackage[algo2e]{algorithm2e} 

\title{Towards Multimodal MIR: Predicting individual differences from music-induced movement}

\multauthor
{Yudhik Agrawal$^1$ \hspace{1cm} Samyak Jain$^1$ \hspace{1cm} Emily Carlson$^2$} { \bfseries{Petri Toiviainen$^2$ \hspace{1cm} Vinoo Alluri$^1$}\\
 $^1$ Cognitive Science Lab, International Institute of Information Technology, Hyderabad, India\\
$^2$ Department of Music, Art and Culture Studies, University of Jyväskylä, Finland\\
{\tt\small \{yudhik.agrawal, samyak.j\}@research.iiit.ac.in,
\{emily.j.carlson, petri.toiviainen\}@jyu.fi}\\ {\tt\small vinoo.alluri@iiit.ac.in}
}

\def\authorname{Y. Agrawal, S. Jain, E. Carlson, P. Toiviainen, and V. Alluri}

\begin{document}

\maketitle
\begin{abstract}
As the field of Music Information Retrieval grows, it is important to take into consideration the multi-modality of music and how aspects of musical engagement such as movement and gesture might be taken into account. Bodily movement is universally associated with music and reflective of important individual features related to music preference such as personality, mood, and empathy. Future multimodal MIR systems may benefit from taking these aspects into account. The current study addresses this by identifying individual differences, specifically Big Five personality traits, and scores on the Empathy and Systemizing Quotients (EQ/SQ) from participants’ free dance movements. Our model successfully explored the unseen space for personality as well as EQ, SQ, which has not previously been accomplished for the latter. $R^{2}$ scores for personality, EQ, and SQ were 76.3\%, 77.1\%, and 86.7\% respectively. As a follow-up, we investigated which bodily joints were most important in defining these traits. We discuss how further research may explore how the mapping of these traits to movement patterns can be used to build a more personalized, multi-modal recommendation system, as well as potential therapeutic applications.
\end{abstract}
\section{Introduction}\label{sec:introduction}
From the perspective of most computational analysis, music can be defined as sound, its important features yielding to the decomposition of waveforms. However, for the vast majority of history, musical sound could not be separated from its source; to whatever degree it may have evolved biologically to serve various human functions, music must be regarded as an embodied and socially embedded phenomenon \cite{bispham2018human, cross2009evolutionary, richter2016don}. Research has shown intimate links between musical features and human movement, including the reflection of hierarchical rhythmic structures in embodied eigen movements \cite{toiviainen2010embodied}, the reflection of higher-level musical structures in group movement to Electronic Dance Movement \cite{solberg2017pleasurable}, and reflection of spectral and timbral features of music in dance \cite{burger2018embodiment}. Bodily movement is one of the most commonly reported responses to music \cite{lesaffre2008potential}, and movement to music is one of very few universal features of music across cultures \cite{nettl2000ethnomusicologist}.

This paper towards Multimodal MIR takes into consideration the multi-modality of music, and takes into account one of the primary aspect of musical engagement, i.e, movement. It is therefore insufficient to consider music only in terms of sound when trying to understand human digital use and interaction with music. This may be especially true in terms of user experience and personalization; human movement in response to music reflects not only the music itself but characteristics of the individual, such as personality \cite{luck2010effects} and emotion \cite{luck2014emotion}. Indeed, research has shown that music-induced movement is so individual that its features can be used in person-identification with a high degree of accuracy \cite{carlson2020dance}. This is in line with previous research, such as that of Cutting et al. \cite{cutting1977recognizing} demonstrating that friends can recognise each other from their walk with only point-light (stick figure) displays of movement, without the need for other distinguishing features. This paradoxical balance between universality and individuality in human motoric responsiveness to music poses a challenge for the creation of digital music interfaces which take music-induced movement into account in providing personalized music experiences.
Although the concept of an interactive music system has long been proposed that allow music playback to be controlled and altered via human gestures \cite{subotnick2001interactive}, human-movement based interaction techniques and devices are fast gaining importance in the field of HCI \cite{gillies2019understanding}. In this context, it makes decoding aspects of a user/individual via human movements a key and useful endeavor, which would then aid in the design of more personalized experiences.

\section{Related Work}
The specific features used in previous work associated movement with individual differences are quite varied. Satchell et al. \cite{satchell2017evidence} examined speed, relative and absolute rotation of the body and found relationships between relative movement of the upper and lower body during walking in both FFM personality traits and gait, while Michalak et al. \cite{michalak2009embodiment} were able to associate low mood with lateral body sway and posture. In dance, relevant features have included amount of movement of the whole body relative to itself and to the environment, responsiveness to music features such as tempo \cite{carlson2016conscientiousness, toiviainen2010embodied}.  Another area for exploration of individual differences in movement patterns has been in the context of disorders that have altered or impaired movement \cite{de2012rehabilitation, anzulewicz2016toward, torres2013autism}.
These links allow us to postulate that movement patterns should give us information related to individual traits and tendencies which can be then linked to music preferences, mood or emotion in relation to music experiences and could have implications for music therapy as well as for music information retrieval. However, as an initial step, there exist no studies that predict personality and empathy as a function of movement patterns. The current study focuses on identifying FFM personality traits, as well as scores on the Empathy Quotient (EQ) and Systemizing Quotient (SQ), from participants’ free dance movements to various genres of music. The EQ measures participants’ tendency to empathize with others \cite{baron2004empathy}, while the SQ measures the tendency to think in terms of systems \cite{baron2003systemizing}. These two measures were originally developed to increase understanding of people with ASD, as in this population
trait systemizing tends to be very high while empathy tends to be low. However previous work has also used the EQ/SQ  to determine how these traits are distributed in the general population. Although previous work has found relationships between empathy and responsiveness to changes in heard music or in dance partner \cite{bamford2019trait, carlson2018dance}, and between EQ/SQ scores and music preferences \cite{carlson2017personality, greenberg2015musical}, general movement patterns associated with empathy have not, to the knowledge of the authors, been explored using dance movement, nor have patterns related to systemizing tendencies.
 % Given the connections between personality and movement, as well as between personality and various music-related settings, it seems reasonable to assume that there are relationships between personality and music-induced movement. 
\section{Method}
\subsection{Participants}
Data acquired was from a previous study \cite{carlson2019empathy} comprising data from 73 university students (54 females, mean age = 25.74 years, std = 4.72 years). Thirty-three reported having received formal musical training; five reported one to three years, ten reported seven to ten years, while sixteen reported ten or more years of training. Seventeen participants reported having received formal dance training; ten reported one to three years, five reported four to six years, while two reported seven to ten years. Participants were of 24 different nationalities, with Finland, the United States, and Vietnam being the most frequently represented. For attending the experiment, participants received two movie ticket vouchers each. All participants spoke and received instructions in English. Fifteen participants were excluded from further analysis due to incomplete data. They were asked to listen to the music and move as freely as they desired, but staying within the marked capture space. The aim of these instructions was to create a naturalistic setting, such that participants would feel free to behave as they might in a real-world situation.

\subsection{Apparatus, Stimuli, and Procedure}
Participants' movements were recorded using a twelve-camera optical motion-capture system (Qualisys Oqus 5+), tracking at a frame rate of 120 Hz, the three-dimensional position of 21 reflective markers attached to each participant. Markers were located as follows (L=left, R=right, F=front, B=back) 1: LF head; 2: RF head; 3: B head; 4: L shoulder; 5: R shoulder; 6: sternum; 7: stomach; 8: LB hip; 9: RB hip; 10: L elbow; 11: R elbow; 12: L wrist; 13: R wrist; 14: L middle finger; 15: R middle finger; 16: L knee; 17: R knee; 18: L ankle; 19: R ankle; 20: L toe; 21: R toe.
The stimuli comprised sixteen 35-second excerpts from eight genres, in randomized order: Blues, Country, Dance, Jazz, Metal, Pop, Rap, and Reggae. 
% \sout{Social tags are defined as ‘free text labels that are applied to items such as artists, albums and songs’ (Lamere, 2008, p. 101), the possibility of which is provided by music-listening platforms such as Last.fm.}
The stimuli for the experiment were selected using a computational process based on social-tagging and acoustic data. The selection pipeline was designed to select naturalistic stimuli that were uncontroversially representative of their respective genres, which would also be appropriate to use in a dance setting. Moreover, investigating movements to multiple genres of music further adds to the generalizability of our findings.
\begin{figure}[h!]
\centering
\begin{subfigure}[b]{.5\linewidth}
  \centering
  \includegraphics[width=\linewidth]{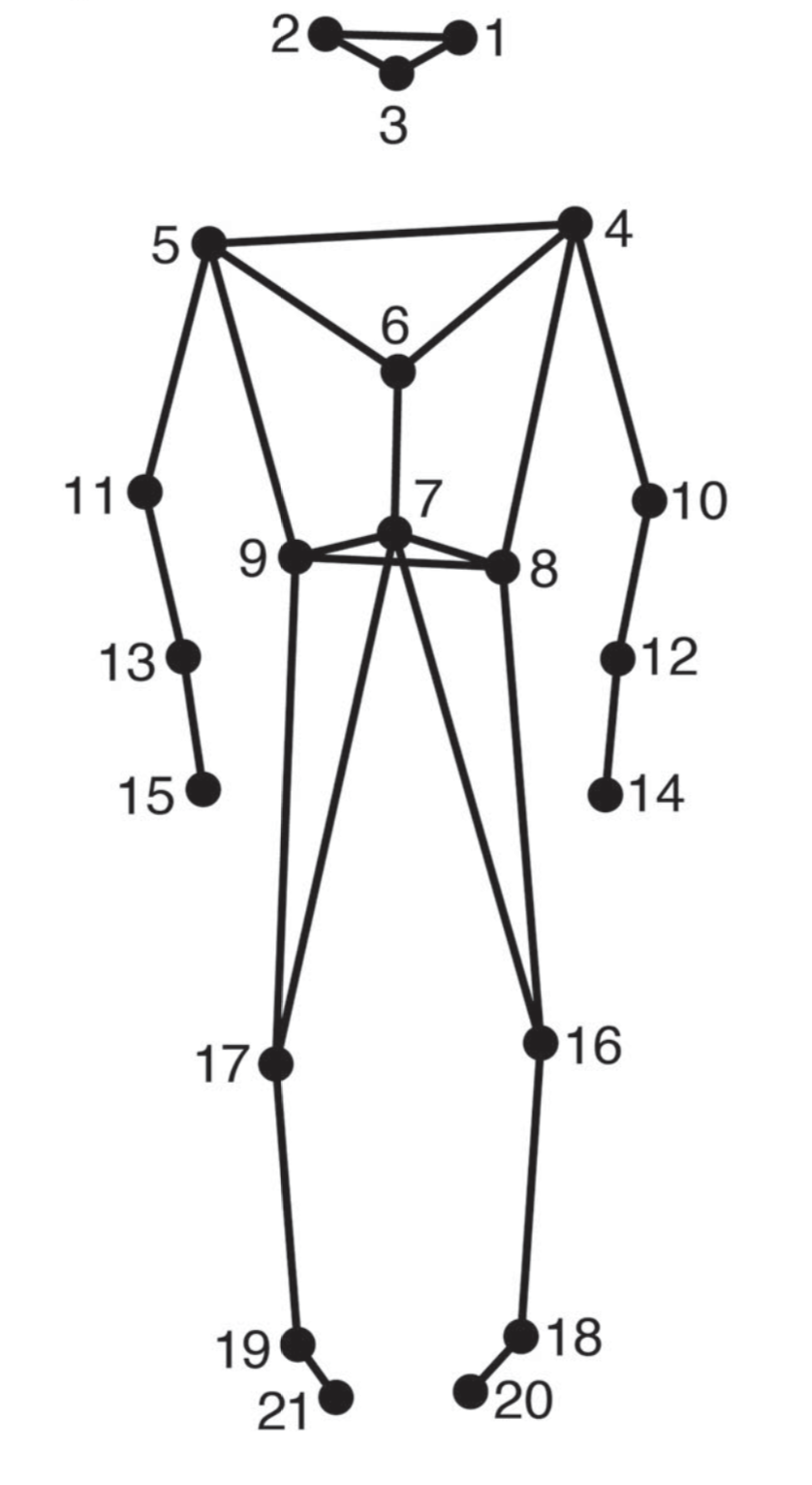}
  \caption*{(A)}
  \label{fig:sub1}
\end{subfigure}%
\begin{subfigure}[b]{.5\linewidth}
  \centering
  \includegraphics[width=.9\linewidth]{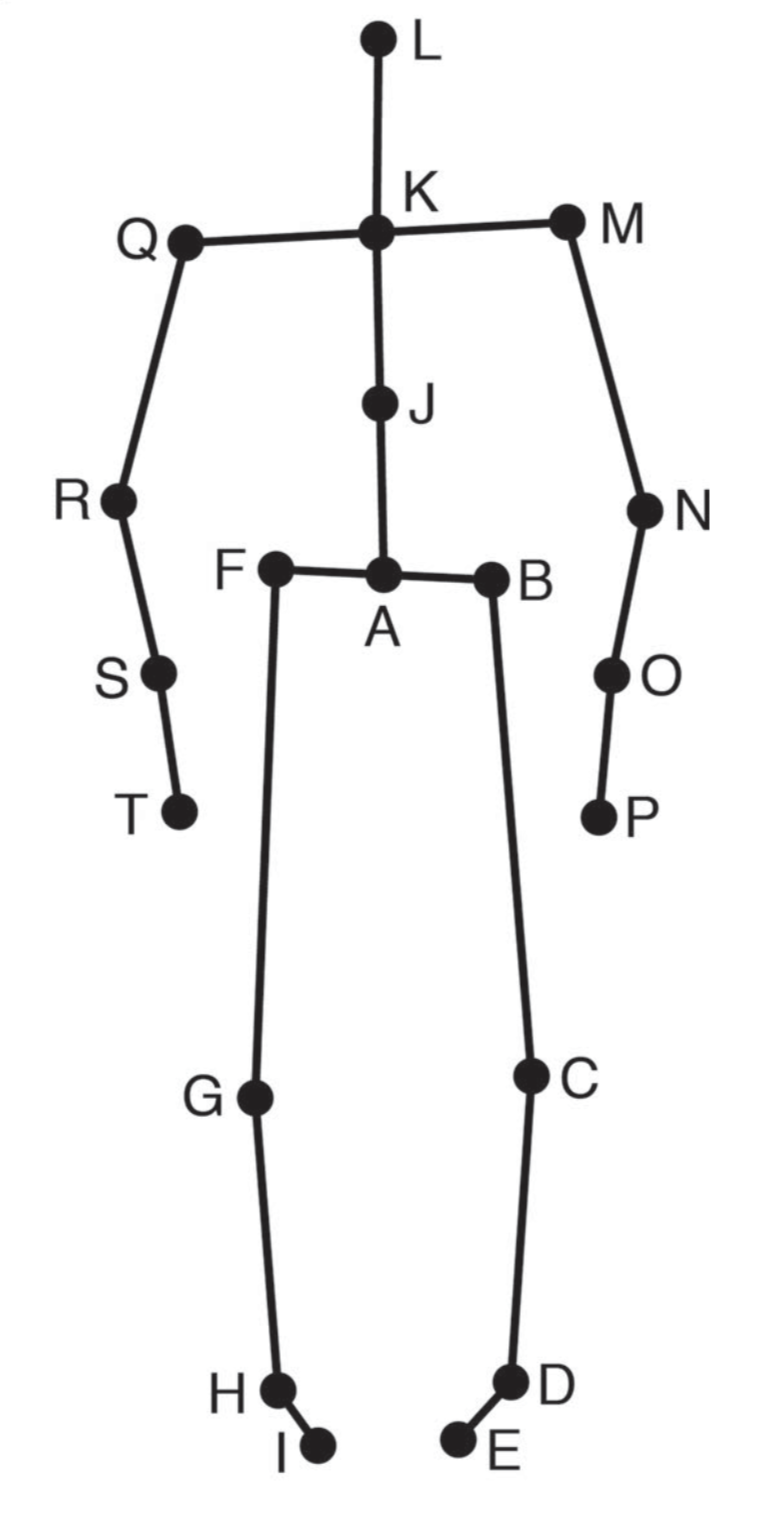}
  \caption*{(B)}
  \label{fig:sub2}
\end{subfigure}
\caption{Marker and joint locations (A) Anterior view of the marker locations a stick figure illustration; (B) Anterior view of the locations of the secondary markers/joints used in animation and analysis of the data}
\label{fig:markers}
\end{figure}
\subsubsection{Personality and Trait Empathy Measures}

The Big Five Inventory (BFI) was used to capture the five predominant personality dimensions, namely, Openness, Conscientiousness, Extraversion, Agreeableness, and Neuroticism \cite{john1999big}.
Trait Empathy was measured using the EQ- and SQ-short form version, developed and validated by Wakabayashi et al. \cite{wakabayashi2006development}, as a result giving an EQ and SQ score per participant. 
\begin{figure*}[t!]
\begin{center}
\includegraphics[height=6.5cm,width=\linewidth]{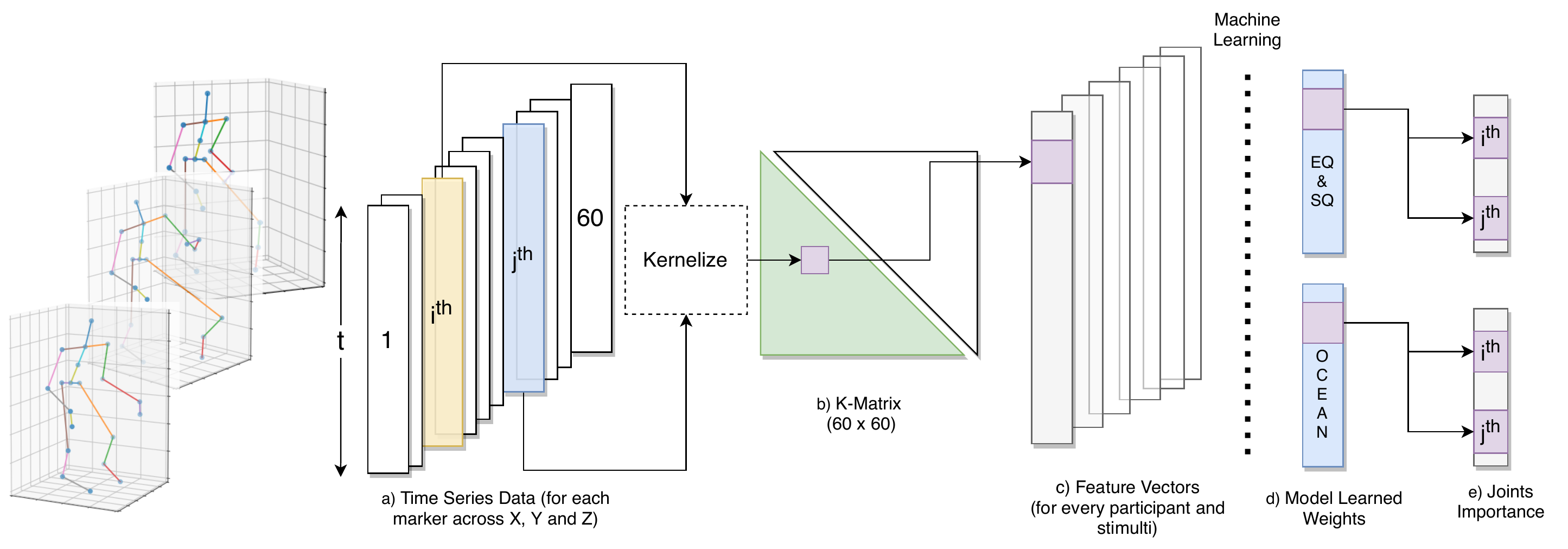}
\caption{Overview of our Pipeline. Given the position of joints across time frames in 3D Euclidean space(a), we apply pairwise correntropy between time series $x_i$ and $x_j$ and calculate the K-matrix (b). Then, taking the lower triangular part of the symmetric covariance matrix, we get the feature vectors (c). After training the regression model on the feature vectors, we get the weight vector(d). Finally, corresponding weight values from the learned weight vector are mapped to the corresponding joints to get the per-joint importance (e).}
\label{fig:JI}
\end{center}
\end{figure*}
\subsection{Feature Extraction}
The analysis and prediction pipeline is illustrated in \figref{fig:JI}. To facilitate extraction of kinematic features using the MATLAB Motion Capture (MoCap) Toolbox\cite{burger2013mocap}, a set of 20 secondary markers, subsequently referred to as joints, was derived from the original 21 markers. The locations of these 20 joints are depicted in \figref{fig:markers}. The locations of joints B, C, D, E, F, G, H, I, M, N, O, P, Q, R, S, and T are identical to the locations of one of the original markers, while the locations of the remaining joints were obtained by averaging the locations of two or more markers; Joint A: midpoint of the two back hip markers; J: midpoint the shoulder and hip markers; K: midpoint of shoulder markers; and L: midpoint of the three head markers. The instantaneous velocity of each marker in each direction was calculated. Instantaneous velocity was estimated by time differentiation followed by the application of a 2nd-order Butterworth filter with a cutoff frequency of 24Hz \cite{burger2013mocap}.

The features used in our analysis is the co-variances of position and velocity. The co-variances between all marker time series in each direction ($X$, $Y$ and $Z$) within each participant for each stimulus. We used a non-linear measure to calculate covariance between the markers. This method, referred to as correntropy between time series $x_{i}$ and $x_{j}$ \cite{liu2007correntropy}, is given by:
\begin{equation}
    K(x_{i}, x_{j}) = e^{\frac{-||x_{i} - x_{j}||_{2}^{2}}{2\sigma^{2}T^2}}
\end{equation}
where $||x_{i} - x_{j}||_{2}$ is L2-norm between $x_{i}$ and $x_{j}$, $\sigma$ is a constant, 12.0 in our case and $T$ is the length of the time-series. The L2-norm is divided by $T$ to normalize according to time series length since it has different lengths with varying stimuli. Since the number of joints are 20 and each joint has three coordinates, the dimension of $K$ would be 60$\times$60. The lower triangular part excluding the diagonal elements of the symmetric covariance matrix was vectorised to produce a feature vector of length 1770 for each participant and for each stimuli.

We also run our experiments using the Normalized feature vectors calculated by using Position and Velocity, we employed standard Gaussian Normalization technique:
\begin{equation}
    \hat{X} = \frac{X - \mu(X)}{\sigma(X)}
\end{equation}
where $\hat{X}$ is the feature vector, $\mu(X)$ is the mean and $\sigma(X)$ is standard deviation.

\subsection{Model Regression}
The most common regression model for value prediction tasks used is Linear Regression. The goal here is to find an optimal line that minimizes the total prediction error. 
% This problem can be modelled as
% \begin{equation}
%     y = W^Tx + b
% \end{equation}{}
% where $x \in R^n$ is the n-dimensional feature vector, $W \in R^n$ is the n-dimensional equation of line, also known as the weight vector and $b$ is bias. $W$ and $b$ are the parameters which should be chosen such that the error is minimized. The error is defined as
% \begin{equation}
%     e = \sum_{i=1}^{n} (y_{i} - \hat{y}_{i})^2
% \end{equation}{}
% where $y_{i}$ is the ground truth value and $\hat{y}_{i}$ is the predicted value. The solution to this problem is
% \begin{equation}
%     \theta = (A^TA)^{-1}A^TY
% \end{equation}{}
% \[
% \theta = 
% \begin{bmatrix}
% W\\
% b
% \end{bmatrix}
% \]
% where $A \in R^{m \times n}$ is the training set and $Y \in R^m$ is the ground truth values. 
But this model treats its parameters as unknown constants whose values must be derived. Moreover, the weights become sensitive when the dataset size is large. So to prevent the model from overfitting, we took principal components of the features to train the model (For the result sections, we will be considering 243 components for position data and 137 components for velocity data which gave us the best results). We also approached this problem by using Bayesian Regression other than Principal Component Regression (PCR)\footnote{Detailed analysis of Principal Component Regression (PCR) and Bayesian Regression are discussed in the supplementary.}.

% So we solved this problem using the Bayesian approach. 

In Bayesian Regression the parameters are treated as random variables belonging to an underlying distribution. Depending on the dataset, we can be more or less certain about the weights. Since, the parameters of the model belong to a distribution, the predictions of the model also belong to a distribution. So we have confidence bounds on our predictions. Therefore, they are better at representing the uncertainty of a model’s predictions. 
% \sout{The model for the Bayesian Regression with the response sampled from a normal distribution is}
% \begin{equation}
%     y \sim N(\beta^Tx, \sigma^2I)
% \end{equation}{}
% \sout{where $\beta$ is the weight vector, $x$ is the feature vector and $\sigma$ is the standard deviation. Here $\beta$ and $\sigma$ are the model parameters. The goal is not to find the single best value of model parameters but to determine the posterior distribution for the model parameters. The posterior probability of the model parameters can be defined as} 
% \begin{equation}
%     p(\beta|D) \propto p(D|\beta)p(\beta)
% \end{equation}
% \begin{equation}
%     \beta \sim N(0, \sigma_{\beta}^2I_{d})    
% \end{equation}

% \sout{where  $p(\beta)$ is the initial probability distribution, also known as prior distribution and $p(D|\beta)$ is known as the likelihood function. Using these approaches, we attempted two tasks 1. EQ and SQ Prediction 2. Personality Prediction.}
\subsubsection{Personality and Trait Empathy Prediction}
The features extracted are used to train five different Bayesian Regression models to predict each of the five personality traits - Openness, Conscientiousness, Extraversion, Agreeableness, and Neuroticism (OCEAN).
% For trait empathy, the range of values of EQ and SQ is 0-80. 
The features extracted are used to train both regression models to predict EQ and SQ respectively. The model is trained and evaluated on the described dataset.

\subsection{Visualizing the Weight Vector}
To interpret the coefficients (also known as the weights or model parameters) of the regression models, we add the value of the feature vector to the corresponding joints. In our algorithm, we first find the index in the 60$\times$60 matrix and then add the absolute value to those joints. 

The sign of the coefficient indicates the direction of the relationship but the magnitude preserves the importance. After that, Min-Max Normalization is applied to bring the values in the range (0, 1) for better visualizing the same variable across similar tasks.
\begin{equation}\label{relativity}
\overline{JI}[i] = \bigg( \frac{JI[i] - min(JI)}{max(JI) - min(JI)} \bigg)  \forall JI[i]
\end{equation}
where $\overline{JI}$ represent the normalized Joint Importance Vector. Algorithm \ref{alg:joint_importance} describes the pseudo-code to get the importance of joints from the weights of the trained regression model.  
\begin{algorithm}[H]
\SetAlgoLined
\KwResult{Calculate a vector $J$ of 20 dimensions representing importance of each joint.}
\hspace{5mm} $W$ is the weight vector; $J$ is the importance vector initialised with $0$; $S$ contains lower triangular indices excluding diagonal indices; 0-indexing is followed;\\

\begin{algorithmic}[1]
\STATE $S\gets LowerTriangularIndices(60 \times 60)$ 
% \STATE $S\gets$empty-list
% \FOR{$i=0:59$}
%     \FOR{$j=0:i-1$}
%         \STATE $S$.push($(i,j)$) 
%     \ENDFOR
% \ENDFOR
\STATE $N\gets$ $S.length()$
\FOR{$k=0:N-1$}
      \STATE $(i,j) := S(k)$
    %   \newline
    % \tcp{mapping the index to the particular joint}
        \STATE $(\hat{i},\hat{j})\gets IndexToJoint(i,j)$
    %   \STATE $i \gets floor(i/3)$, $j \gets floor(j/3)$ 
      \STATE $J(\hat{i}):= J(\hat{i}) + |W(k)|$
      \STATE $J(\hat{j}):= J(\hat{j}) + |W(k)|$
\ENDFOR
\RETURN $J$
\end{algorithmic}

\caption{Joint Importance}
\label{alg:joint_importance}
\end{algorithm}
After getting a vector of 20 dimension, we reduce it to 12 before visualizing joint importance. We did this by taking the average of joints which occur in pairs eg. (L shoulder, R shoulder), (L wrist, R wrist).

%\section{Experiments}

\subsubsection{Evaluation Metric}
(a) Root Mean Square Error (RMSE): It computes a risk metric corresponding to the expected value of the root of squared (quadratic) error or loss. 
% \sout{If $\hat{y}_{i}$ is the predicted value of the $i^{th}$ sample and $y_i$ is the corresponding true value for total $n$ samples, then the RMSE estimated is defined as:}
% \begin{equation}\label{relativity}
% RMSE(y, \hat{y}) =  \sqrt{\frac{1}{n} \sum_{i=1}^{n}(y_i - \hat{y}_i)^2}
% \end{equation}
\\
(b) $R^{2}$ Score: It represents the proportion of the variance(of y) that has been explained by the independent variables in the model.\footnote{Detailed explanation of metrics is provided in the supplementary.} 
% \sout{If $\hat{y}_{i}$ is the predicted value of the $i^{th}$ sample, $y_i$ is the corresponding true value for total $n$ samples, and $\bar{y}$ is the mean of the ground truth data, the estimated $R^2$ is defined as:}
% \begin{equation}\label{relativity}
% R^2(y, \hat{y}) =  1 -  \frac{\sum_{i=1}^{n} (y_i - \hat{y}_i)^2}{\sum_{i=1}^{n}(y_i -\bar{y})^2}
% \end{equation}
\\
As the square root of a variance, RMSE can be interpreted as the standard deviation of the unexplained variance, and has the useful property of being in the same units as the response variable and at the same time the $R^2$ helps us evaluate the goodness of fit in capturing the variance in training data. We calculate RMSE and $R^2$ on multiple splits so that we get an average estimate of the accuracy.
%For this reason, it is often useful to review and comment on both.

\subsection{Results}
\subsubsection{EQ and SQ}
The results for EQ prediction are in \tabref{table:eq_results} and SQ prediction are in \tabref{table:sq_results}. The results are calculated using 5-fold cross validation. The range of EQ and SQ is 0-80. The boldface values represent the best score. The 'N' in the tables denote that Gaussian Normalization was applied on the features. We trained and evaluated two different models for each of the aforementioned tasks. We can see that using position data, instead of velocity data, to generate the feature vectors, gave us the best results. Also, we can see that the Bayesian Regression gave better results than Principal Component Regression on both metrics. So from here on, we will be using Bayesian Regression for other prediction and analysis tasks. \\
%\textbf{Gender should not be the defining parameter for discussing EQ-SQ as moving towards a progressing society, the span of this social constructs is more continuous than that which prevailed in previous research.}
\begin{table}[h!]
\centering
\begin{tabular}{|c|c c|c c|}
\hline
Input & \multicolumn{2}{c|}{PCR} & \multicolumn{2}{c|}{Bayesian  Ridge} \\
 & RMSE & $R^{2}$ & RMSE & $R^{2}$ \\ \hline
\textbf{Position} & 3.071 & 0.708 & \textbf{2.722} & \textbf{0.771} \\
\begin{tabular}[c]{@{}c@{}}\textbf{Position(N)}\end{tabular} & 3.201 & 0.684 & 2.733 & 0.765 \\
\textbf{Velocity} & 4.938 & 0.249 & 4.343 & 0.423 \\
\begin{tabular}[c]{@{}c@{}}\textbf{Velocity(N)}\end{tabular} & 4.583 & 0.353 & 4.015 & 0.503 \\ \hline
\end{tabular}
\caption{Prediction Results for Empathizing Quotient}
\label{table:eq_results}
\end{table}

\begin{table}[h!]
\centering
\begin{tabular}{|c|c c|c c|}
\hline
Input & \multicolumn{2}{c|}{PCR} & \multicolumn{2}{c|}{Bayesian  Ridge} \\
 & RMSE & $R^{2}$ & RMSE & $R^{2}$ \\ \hline
\textbf{Position} & 2.398 & 0.781 & \textbf{2.161} & \textbf{0.867} \\
\begin{tabular}[c]{@{}c@{}}\textbf{Position(N)}\end{tabular} & 2.363 & 0.786 & 2.502 & 0.838 \\
\textbf{Velocity} & 4.448 & 0.252 & 3.832 & 0.469 \\
\begin{tabular}[c]{@{}c@{}}\textbf{Velocity(N)}\end{tabular} & 4.211 & 0.329 & 3.714 & 0.552 \\ \hline
\end{tabular}
\caption{Prediction Results for Systemizing Quotient}
\label{table:sq_results}
\end{table}

\begin{table*}[t!]
\centering
\begin{tabular}{|c|c c|c c|c c|c c|c c|}
\hline
Input & \multicolumn{2}{c|}{Openness} & \multicolumn{2}{c|}{Conscientiousness} & \multicolumn{2}{c|}{Extraversion} & \multicolumn{2}{c|}{Agreeableness} & \multicolumn{2}{c|}{Neuroticism}\\
 & RMSE & $R^{2}$ & RMSE & $R^{2}$ & RMSE & $R^{2}$ & RMSE & $R^{2}$ & RMSE & $R^{2}$ \\ \hline
\textbf{Position} & \textbf{0.197} & \textbf{0.776} & \textbf{0.317} & \textbf{0.760} & \textbf{0.384} & 0.743 & \textbf{0.252} & \textbf{0.776} & \textbf{0.384} & \textbf{0.758} \\
\begin{tabular}[c]{@{}c@{}} \textbf{Position(N)}\end{tabular} & 0.227 & 0.740 & 0.332 & 0.690 & 0.414 & \textbf{0.756} & 0.273 & 0.716 & 0.390 & 0.739  \\
\textbf{Velocity} & 0.332 & 0.464 & 0.487 & 0.415 & 0.556 & 0.523 & 0.440 & 0.335 & 0.557 & 0.483  \\
\begin{tabular}[c]{@{}c@{}} \textbf{Velocity(N)}\end{tabular} & 0.304 & 0.527 & 0.426 & 0.543 & 0.501 & 0.623 & 0.408 & 0.442 & 0.461 & 0.654 \\ 
\hline
\end{tabular}
\caption{Prediction Results for Five Personality Traits using Bayesian Regression}
\label{table:ocean_results_d2}
\end{table*}

% \subsubsection{Correlation between BFI and TIPI}
% \tabref{table:ocean_corr} shows the correlations between the two different Personality Models namely BFI and TIPI. Pearson correlations revealed significant positive correlation between different methods of Personality Trait evaluations.
% All the correlations with significance values less than p<.01 are indicated in boldface.

% \begin{table}[h!]
% \begin{tabular}{|c|ccccc|}
% \hline
%  & O(TIPI) & C(TIPI) & E(TIPI) & A(TIPI) & N(TIPI) \\
%  \hline
% O(BFI) & \textbf{0.571**} & -0.239 & -0.062 & -0.035 & 0.292 \\
% C(BFI) & -0.032 & \textbf{0.495**} & 0.102 & 0.291 & -0.246 \\
% E(BFI) & 0.261 & 0.058 & \textbf{0.821**} & 0.042 & -0.178 \\
% A(BFI) & 0.061 & 0.097 & \textbf{0.337} & \textbf{0.396*} & -0.098 \\
% N(BFI) & -0.018 & -0.147 & -0.262 & -0.243 & \textbf{0.717**}\\ \hline
% \end{tabular}
% *p<0.003, **p < 0.0001
% \caption{Results of the Pearson correlation between the OCEAN values calculated through BFI and TIPI.}
% \label{table:ocean_corr}
% \end{table}

\subsubsection{Personality Regression}
The results for OCEAN value prediction on Dataset can be found in \tabref{table:ocean_results_d2}. The results are calculated using 5-fold cross validation. The range of the personality values is 1.0-5.0. We can see that using position data to extract features gave the best results on predicting all five personality traits on the dataset. We can concur that using position data instead of velocity data in the kernelized space is better for these regression tasks.  

\subsubsection{Joints' Importance}
For evaluating joint importance we used learned weights of the model using position data across different tasks. For the purpose of analyzing the importance of the joints, we reduced them to 12 by taking the average for those joints which occur in pairs eg. (L shoulder, R shoulder). This was also done for hips, knee, ankle, toe, elbow, wrist, and finger. 
% In general, we can say that the farther away from the mean the joint importance value for an individual joint is, the more important it is in characterizing that trait. 
Altogether the results in characterizing an individual trait is dominated by the limbs than the core of the body.

\begin{figure}[b!]
\includegraphics[width=\linewidth]{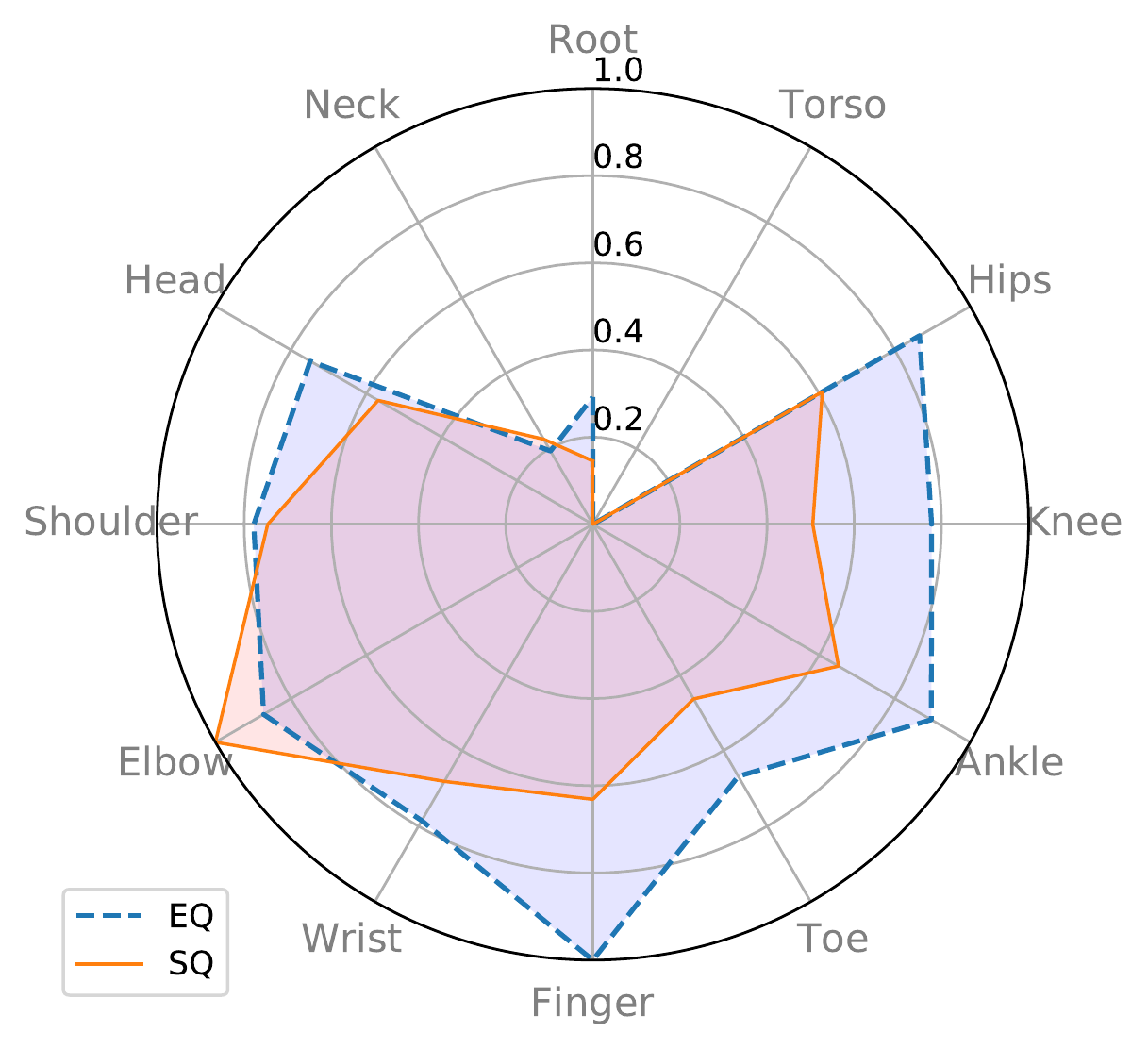}
\caption{Relative importance of Joints in EQ and SQ Tasks using the Position Data. }
\label{fig:eq_sq_spyder}
\end{figure}

From the relative joint importance depicted in \figref{fig:eq_sq_spyder}, we observe that 'Ankle', Elbow' and 'Shoulder' play an important role in determining EQ and SQ of an individual, whereas 'Neck' and 'Torso' have a negligible contribution. We also infer that 'Finger', 'Hip', and 'Knee' are more crucial joints for predicting EQ than for SQ whereas 'Elbow' holds significantly higher importance for predicting SQ than for EQ.

\begin{figure*}[ht!]
\begin{center}
\includegraphics[width=\linewidth]{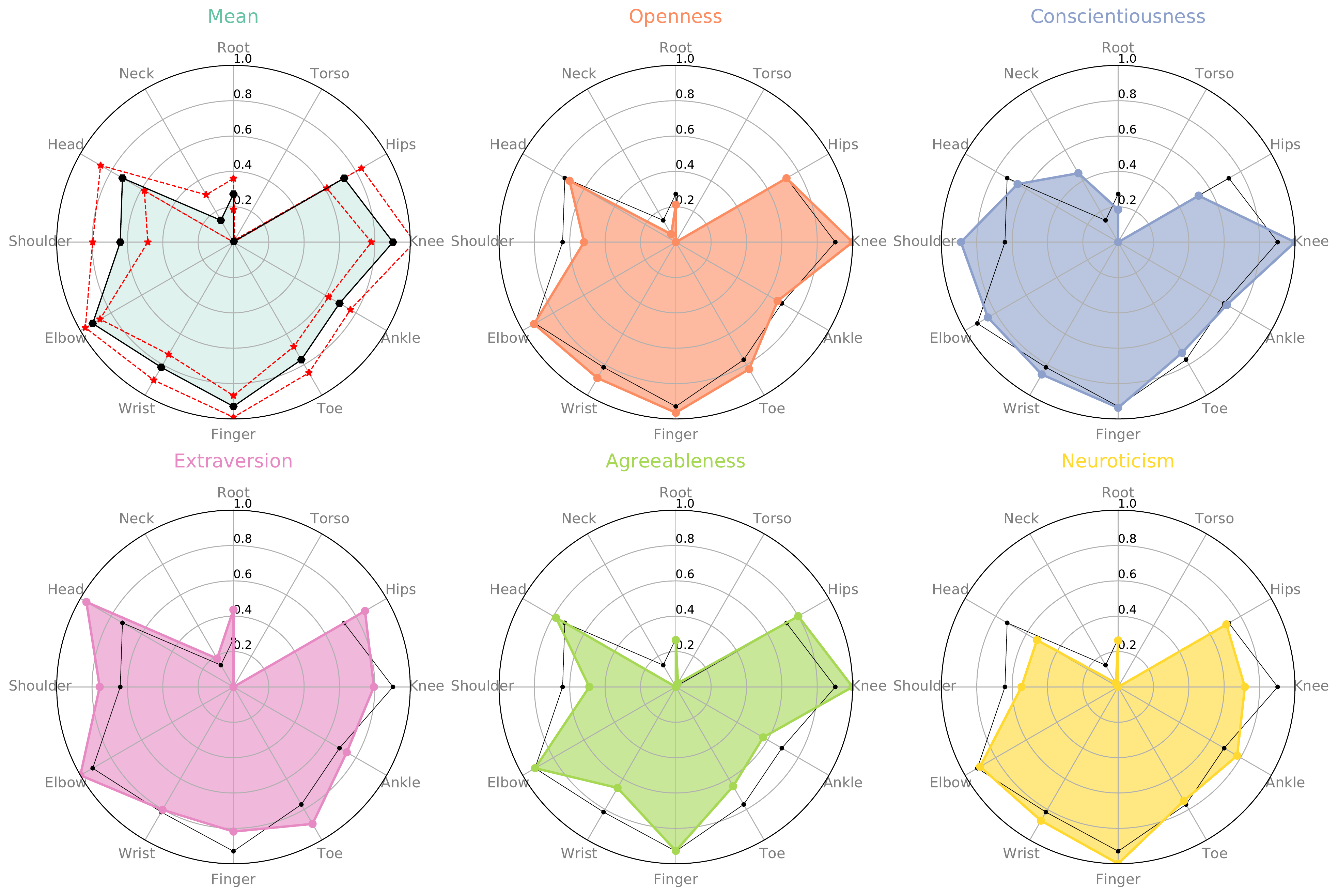}
\caption{Relative importance of Joints of the five personality traits(Openness,  Conscientiousness,  Extraversion, Agreeableness, and Neuroticism) using the Position Data. The black line indicates the mean importance of the corresponding joint marker. The red dotted line in the top left sub-figure indicates the standard deviation about the mean. }
\label{fig:ocean_spyder}
\end{center}
\end{figure*}

\figref{fig:ocean_spyder} displays the relative joint importance of personality along with the mean plotted in each sub-figure. 
The farther away from the mean the joint importance value for an individual joint is, the more important it is in characterizing that trait. 
Some similarities in the joint importance profiles across the personality traits can be attributed to the inherent correlation that exists among them\footnote{The table for Spearman Correlation between the personality traits is provided in the supplementary material.}.
We observe that it is the 'Finger', 'Elbow', and 'Knees' that contribute to Feature Importance whereas 'Root', 'Neck' and 'Torso' have negligible contribution. For characterizing Conscientiousness, 'Shoulders', 'Knees' and 'Neck' play a crucial role while 'Head' and 'Toe' plays an important role for Extraversion. For Agreeableness, 'Neck' and 'Wrists' have relatively less importance as compared to other  joints whereas, 'Wrists' play an important role in Openness. Finally, there are no significant defining features for Neuroticism, which indicates that their expression in Dance Movements through Music-Induced Movements are very limited.

% Altogether the results in characterizing an individual trait have more to do with the limbs than that with the core of the body. 
% Some similarities in the joint importance which we observe between the personality can be justified from the Spearman Correlation table provided in the supplementary material.
% For EQ, 'Finger', 'Ankle' and 'Shoulder' play a crucial role. For SQ, 'Elbow', 'Shoulder' and 'Ankle' play an important role. 

\section{Discussion}
 Music experiences are highly embodied, making it necessary to consider individual embodied responses to music in developing more advanced personalized user experiences. The current study is among the first to the authors’ knowledge to use participants’ free dance movements to predict personality traits, and both the Empathizing and Systemizing Quotients (EQ/SQ). 
 
Co-variance between joint velocities has previously been used to identify an individual from their free dance movements with a high degree of accuracy \cite{carlson2020dance}. The results of the current analysis show co-variance to be a useful feature in predicting individual differences. However, we achieved considerably better prediction accuracy by using position data than velocity data. 

Overall, the limbs of the body seemed to have more importance in predicting individual traits than the core body. This is in line with the fact that gesture plays an important role in communication \cite{goldin2006talking}, and as specifically regards the EQ/SQ, as these tests were originally developed in conjunction with studies of ASD, in which gesture and imitative movement appear to be compromised \cite{hamilton2007imitation}. Although the sample used in the current study comprised typically functioning (non-ASD) participants, the accuracy of prediction of EQ/SQ scores in this analysis is worth highlighting in light of recent work suggesting the existence of motor signatures unique to ASD, detectable from whole body movements as well as data drawn from participants’ interaction with tablets \cite{anzulewicz2016toward,  wu2018biomarker}.

The specific markers that were important in the prediction of individual traits in some cases corroborates previous work, and in some cases contradicts it. Luck et al.\cite{luck2010effects} found correlations between Extraversion and speed of head movement, which supports the current finding that the head is of particular importance in identifying Extraversion. Carlson et al. \cite{carlson2016conscientiousness} found that, compared to Conscientiousness, the core body was more important in responsiveness to musical tempo in relation to Extraversion, which is partly supported by the slightly greater importance of the finger and wrist markers to Extraversion in our study, but partly contradicted by the importance of the shoulder marker in Extraversion. The difference between findings may relate to the use in the current study of position rather than velocity or acceleration data; that is, core body posture while moving to music may be more indicative of Conscientiousness than core body movement. EQ scores were more related to head, finger, hips and lower limb joints than SQ scores, which may be partly attributed to gender-typical movement patterns as females tend to score higher on the EQ than males \cite{baron2004empathy, troje2002decomposing}.

Several limitations of the current study should be noted. First, the majority of participants were from European or North American countries, and all eight music stimuli were of Western origin, limiting the degree to which results can be generalized cross-culturally. Secondly, There may exist potential bias due to gender imbalance. Future work could include separate analysis performed within gender categories. And lastly, participants' preferences for heard stimuli were not included in our model. This would be an important feature to focus on in future work, as preference and enjoyment are highly relevant to personalized MIR.

Further extension of this work could help to make music recommendation systems more robust. Previous work has considered the relationship between personality and music preference \cite{carlson2017personality, rentfrow2003re}, while Greenberg et al. \cite{greenberg2015musical} explored the relationship between music preference and empathizing-systematizing theory, suggesting even that music may play a role in increasing empathy in people with empathy-related disorders, such as ASD. However, the relationship between embodiment, personality and musical experiences requires further exploration.

To conclude, this study represents an early step towards multimodal MIR.  To make this approach applicable to personalized gesture-based retrieval systems, it can be extended to monocular video captured by accessible devices such as a mobile phone camera. This approach would be feasible due to recent progress in the area of 3D human pose estimation in predicting the body joint coordinates from a monocular video \cite{pavllo20193d, venkat2019humanmeshnet, cheng20203d}. This would then allow future recommendation systems to take embodied processes into account, resulting in better and more responsive personalized experiences.

\clearpage
% For bibtex users:
\bibliography{ISMIRtemplate}

\begin{thebibliography}{10}
\providecommand{\url}[1]{#1}
\csname url@samestyle\endcsname
\providecommand{\newblock}{\relax}
\providecommand{\bibinfo}[2]{#2}
\providecommand{\BIBentrySTDinterwordspacing}{\spaceskip=0pt\relax}
\providecommand{\BIBentryALTinterwordstretchfactor}{4}
\providecommand{\BIBentryALTinterwordspacing}{\spaceskip=\fontdimen2\font plus
\BIBentryALTinterwordstretchfactor\fontdimen3\font minus
  \fontdimen4\font\relax}
\providecommand{\BIBforeignlanguage}[2]{{%
\expandafter\ifx\csname l@#1\endcsname\relax
\typeout{** WARNING: IEEEtran.bst: No hyphenation pattern has been}%
\typeout{** loaded for the language `#1'. Using the pattern for}%
\typeout{** the default language instead.}%
\else
\language=\csname l@#1\endcsname
\fi
#2}}
\providecommand{\BIBdecl}{\relax}
\BIBdecl

\bibitem{bispham2018human}
J.~C. Bispham, ``The human faculty for music: What's special about it?'' Ph.D.
  dissertation, University of Cambridge, 2018.

\bibitem{cross2009evolutionary}
I.~Cross, ``The evolutionary nature of musical meaning,'' \emph{Musicae
  scientiae}, vol.~13, no. 2\_suppl, pp. 179--200, 2009.

\bibitem{richter2016don}
J.~Richter and R.~Ostovar, ``“it don’t mean a thing if it ain’t got that
  swing”--an alternative concept for understanding the evolution of dance and
  music in human beings,'' \emph{Frontiers in human neuroscience}, vol.~10, p.
  485, 2016.

\bibitem{toiviainen2010embodied}
P.~Toiviainen, G.~Luck, and M.~R. Thompson, ``Embodied meter: hierarchical
  eigenmodes in music-induced movement,'' \emph{Music Perception}, vol.~28,
  no.~1, pp. 59--70, 2010.

\bibitem{solberg2017pleasurable}
R.~T. Solberg and A.~R. Jensenius, ``Pleasurable and intersubjectively embodied
  experiences of electronic dance music,'' \emph{Empirical Musicology Review},
  vol.~11, no. 3-4, pp. 301--318, 2017.

\bibitem{burger2018embodiment}
B.~Burger and P.~Toiviainen, ``Embodiment in electronic dance music: Effects of
  musical content and structure on body movement,'' \emph{Musicae Scientiae},
  p. 1029864918792594, 2018.

\bibitem{lesaffre2008potential}
M.~Lesaffre, L.~D. Voogdt, M.~Leman, B.~D. Baets, H.~D. Meyer, and J.-P.
  Martens, ``How potential users of music search and retrieval systems describe
  the semantic quality of music,'' \emph{Journal of the American Society for
  Information Science and Technology}, vol.~59, no.~5, pp. 695--707, 2008.

\bibitem{nettl2000ethnomusicologist}
B.~Nettl, ``An ethnomusicologist contemplates universals in musical sound and
  musical culture,'' \emph{The origins of music}, vol.~3, no.~2, pp. 463--472,
  2000.

\bibitem{luck2010effects}
G.~Luck, S.~Saarikallio, B.~Burger, M.~R. Thompson, and P.~Toiviainen,
  ``Effects of the big five and musical genre on music-induced movement,''
  \emph{Journal of Research in Personality}, vol.~44, no.~6, pp. 714--720,
  2010.

\bibitem{luck2014emotion}
G.~Luck, S.~Saarikallio, B.~Burger, M.~Thompson, and P.~Toiviainen,
  ``Emotion-driven encoding of music preference and personality in dance,''
  \emph{Musicae Scientiae}, vol.~18, no.~3, pp. 307--323, 2014.

\bibitem{carlson2020dance}
E.~Carlson, P.~Saari, B.~Burger, and P.~Toiviainen, ``Dance to your own drum:
  Identification of musical genre and individual dancer from motion capture
  using machine learning,'' \emph{Journal of New Music Research}, pp. 1--16,
  2020.

\bibitem{cutting1977recognizing}
J.~E. Cutting and L.~T. Kozlowski, ``Recognizing friends by their walk: Gait
  perception without familiarity cues,'' \emph{Bulletin of the psychonomic
  society}, vol.~9, no.~5, pp. 353--356, 1977.

\bibitem{subotnick2001interactive}
M.~Subotnick, ``Interactive music playback system utilizing gestures,'' Nov.~1
  2001, uS Patent App. 09/835,840.

\bibitem{gillies2019understanding}
M.~Gillies, ``Understanding the role of interactive machine learning in
  movement interaction design,'' \emph{ACM Transactions on Computer-Human
  Interaction (TOCHI)}, vol.~26, no.~1, pp. 1--34, 2019.

\bibitem{satchell2017evidence}
L.~Satchell, P.~Morris, C.~Mills, L.~O’Reilly, P.~Marshman, and L.~Akehurst,
  ``Evidence of big five and aggressive personalities in gait biomechanics,''
  \emph{Journal of nonverbal behavior}, vol.~41, no.~1, pp. 35--44, 2017.

\bibitem{michalak2009embodiment}
J.~Michalak, N.~F. Troje, J.~Fischer, P.~Vollmar, T.~Heidenreich, and
  D.~Schulte, ``Embodiment of sadness and depression—gait patterns associated
  with dysphoric mood,'' \emph{Psychosomatic medicine}, vol.~71, no.~5, pp.
  580--587, 2009.

\bibitem{carlson2016conscientiousness}
E.~Carlson, B.~Burger, J.~London, M.~R. Thompson, and P.~Toiviainen,
  ``Conscientiousness and extraversion relate to responsiveness to tempo in
  dance,'' \emph{Human movement science}, vol.~49, pp. 315--325, 2016.

\bibitem{de2012rehabilitation}
M.~J. de~Dreu, A.~Van Der~Wilk, E.~Poppe, G.~Kwakkel, and E.~E. van Wegen,
  ``Rehabilitation, exercise therapy and music in patients with parkinson's
  disease: a meta-analysis of the effects of music-based movement therapy on
  walking ability, balance and quality of life,'' \emph{Parkinsonism \& related
  disorders}, vol.~18, pp. S114--S119, 2012.

\bibitem{anzulewicz2016toward}
A.~Anzulewicz, K.~Sobota, and J.~T. Delafield-Butt, ``Toward the autism motor
  signature: Gesture patterns during smart tablet gameplay identify children
  with autism,'' \emph{Scientific reports}, vol.~6, no.~1, pp. 1--13, 2016.

\bibitem{torres2013autism}
E.~B. Torres, M.~Brincker, R.~W. Isenhower~III, P.~Yanovich, K.~A. Stigler,
  J.~I. Nurnberger~Jr, D.~N. Metaxas, and J.~V. Jos{\'e}, ``Autism: the
  micro-movement perspective,'' \emph{Frontiers in integrative neuroscience},
  vol.~7, p.~32, 2013.

\bibitem{baron2004empathy}
S.~Baron-Cohen and S.~Wheelwright, ``The empathy quotient: an investigation of
  adults with asperger syndrome or high functioning autism, and normal sex
  differences,'' \emph{Journal of autism and developmental disorders}, vol.~34,
  no.~2, pp. 163--175, 2004.

\bibitem{baron2003systemizing}
S.~Baron-Cohen, J.~Richler, D.~Bisarya, N.~Gurunathan, and S.~Wheelwright,
  ``The systemizing quotient: an investigation of adults with asperger syndrome
  or high--functioning autism, and normal sex differences,''
  \emph{Philosophical Transactions of the Royal Society of London. Series B:
  Biological Sciences}, vol. 358, no. 1430, pp. 361--374, 2003.

\bibitem{bamford2019trait}
J.~M.~S. Bamford and J.~W. Davidson, ``Trait empathy associated with
  agreeableness and rhythmic entrainment in a spontaneous movement to music
  task: Preliminary exploratory investigations,'' \emph{Musicae Scientiae},
  vol.~23, no.~1, pp. 5--24, 2019.

\bibitem{carlson2018dance}
E.~Carlson, B.~Burger, and P.~Toiviainen, ``Dance like someone is watching: A
  social relations model study of music-induced movement,'' \emph{Music \&
  Science}, vol.~1, p. 2059204318807846, 2018.

\bibitem{carlson2017personality}
E.~Carlson, P.~Saari, B.~Burger, and P.~Toiviainen, ``Personality and musical
  preference using social-tagging in excerpt-selection.''
  \emph{Psychomusicology: Music, Mind, and Brain}, vol.~27, no.~3, p. 203,
  2017.

\bibitem{greenberg2015musical}
D.~M. Greenberg, S.~Baron-Cohen, D.~J. Stillwell, M.~Kosinski, and P.~J.
  Rentfrow, ``Musical preferences are linked to cognitive styles,'' \emph{PloS
  one}, vol.~10, no.~7, 2015.

\bibitem{carlson2019empathy}
E.~Carlson, B.~Burger, and P.~Toiviainen, ``Empathy, entrainment, and perceived
  interaction in complex dyadic dance movement,'' \emph{Music Perception: An
  Interdisciplinary Journal}, vol.~36, no.~4, pp. 390--405, 2019.

\bibitem{john1999big}
O.~P. John, S.~Srivastava \emph{et~al.}, ``The big five trait taxonomy:
  History, measurement, and theoretical perspectives,'' \emph{Handbook of
  personality: Theory and research}, vol.~2, no. 1999, pp. 102--138, 1999.

\bibitem{wakabayashi2006development}
A.~Wakabayashi, S.~Baron-Cohen, S.~Wheelwright, N.~Goldenfeld, J.~Delaney,
  D.~Fine, R.~Smith, and L.~Weil, ``Development of short forms of the empathy
  quotient (eq-short) and the systemizing quotient (sq-short),''
  \emph{Personality and individual differences}, vol.~41, no.~5, pp. 929--940,
  2006.

\bibitem{burger2013mocap}
B.~Burger and P.~Toiviainen, ``Mocap toolbox-a matlab toolbox for computational
  analysis of movement data,'' in \emph{10th Sound and Music Computing
  Conference, SMC 2013, Stockholm, Sweden}.\hskip 1em plus 0.5em minus
  0.4em\relax Logos Verlag Berlin, 2013.

\bibitem{liu2007correntropy}
W.~Liu, P.~P. Pokharel, and J.~C. Pr{\'\i}ncipe, ``Correntropy: Properties and
  applications in non-gaussian signal processing,'' \emph{IEEE Transactions on
  Signal Processing}, vol.~55, no.~11, pp. 5286--5298, 2007.

\bibitem{goldin2006talking}
S.~Goldin-Meadow, ``Talking and thinking with our hands,'' \emph{Current
  directions in psychological science}, vol.~15, no.~1, pp. 34--39, 2006.

\bibitem{hamilton2007imitation}
A.~F. d.~C. Hamilton, R.~M. Brindley, and U.~Frith, ``Imitation and action
  understanding in autistic spectrum disorders: how valid is the hypothesis of
  a deficit in the mirror neuron system?'' \emph{Neuropsychologia}, vol.~45,
  no.~8, pp. 1859--1868, 2007.

\bibitem{wu2018biomarker}
D.~Wu, J.~V. Jos{\'e}, J.~I. Nurnberger, and E.~B. Torres, ``A biomarker
  characterizing neurodevelopment with applications in autism,''
  \emph{Scientific reports}, vol.~8, no.~1, pp. 1--14, 2018.

\bibitem{troje2002decomposing}
N.~F. Troje, ``Decomposing biological motion: A framework for analysis and
  synthesis of human gait patterns,'' \emph{Journal of vision}, vol.~2, no.~5,
  pp. 2--2, 2002.

\bibitem{rentfrow2003re}
P.~J. Rentfrow and S.~D. Gosling, ``The do re mi's of everyday life: the
  structure and personality correlates of music preferences.'' \emph{Journal of
  personality and social psychology}, vol.~84, no.~6, p. 1236, 2003.

\bibitem{pavllo20193d}
D.~Pavllo, C.~Feichtenhofer, D.~Grangier, and M.~Auli, ``3d human pose
  estimation in video with temporal convolutions and semi-supervised
  training,'' in \emph{Proceedings of the IEEE Conference on Computer Vision
  and Pattern Recognition}, 2019, pp. 7753--7762.

\bibitem{venkat2019humanmeshnet}
A.~Venkat, C.~Patel, Y.~Agrawal, and A.~Sharma, ``Humanmeshnet: Polygonal mesh
  recovery of humans,'' in \emph{Proceedings of the IEEE International
  Conference on Computer Vision Workshops}, 2019, pp. 0--0.

\bibitem{cheng20203d}
Y.~Cheng, B.~Yang, B.~Wang, and R.~T. Tan, ``3d human pose estimation using
  spatio-temporal networks with explicit occlusion training,'' \emph{arXiv
  preprint arXiv:2004.11822}, 2020.

\end{thebibliography}

\end{document}

%
% \maketitle
%
% \begin{center}

\twocolumn[ \section*{Supplementary Material of "Towards Multimodal MIR: Predicting individual differences from music-induced movement"}]
% \end{center}

\noindent In this supplementary material, we present more experimental results and details about the Evaluation Metric and Bayesian Regression that could not be included in the main manuscript due to the lack of space.\\

\subsection*{Evaluation Metrics}

(a) Root-Mean Square Error (RMSE): If $\hat{y}_{i}$ is the predicted value of the $i^{th}$ sample and $y_i$ is the corresponding true value for total $n$ samples, then the RMSE estimated is defined as:
\begin{equation}\label{relativity}
RMSE(y, \hat{y}) =  \sqrt{\frac{1}{n} \sum_{i=1}^{n}(y_i - \hat{y}_i)^2}
\end{equation}
We can interpret the model's performance using RMSE. Consider $X$ being the ground truth value and $Y$ is the RMSE score, then we can say that the model's prediction value will be accurate in the range $X-Y$ to $X+Y$.\newline \\
(b) $R^{2}$ Score: If $\hat{y}_{i}$ is the predicted value of the $i^{th}$ sample, $y_i$ is the corresponding true value for total $n$ samples, and $\bar{y}$ is the mean of the ground truth data, the estimated $R^2$ is defined as:
\begin{equation}\label{relativity}
R^2(y, \hat{y}) =  1 -  \frac{\sum_{i=1}^{n} (y_i - \hat{y}_i)^2}{\sum_{i=1}^{n}(y_i -\bar{y})^2}
\end{equation}

\subsection*{Principal Component Regression}
The Linear Regression model was used to predict the EQ and SQ values but we found that the model was highly overfitting.
% as it was giving as the $R^2$ score for training and testing set came out to be 1.0 and -19.0 respectively. This was due to the number of features being more than the number of data points.
So, we took Principal Components of the features for this model. We calculated the RMSE and $R^2$ scores for both of the aforementioned tasks after taking the Principal Components. We repeated this experiment by varying the number of principal components.

\subsubsection*{EQ}
\begin{figure}[h!]
  \begin{subfigure}[b]{0.49\linewidth}
    \includegraphics[width=\linewidth]{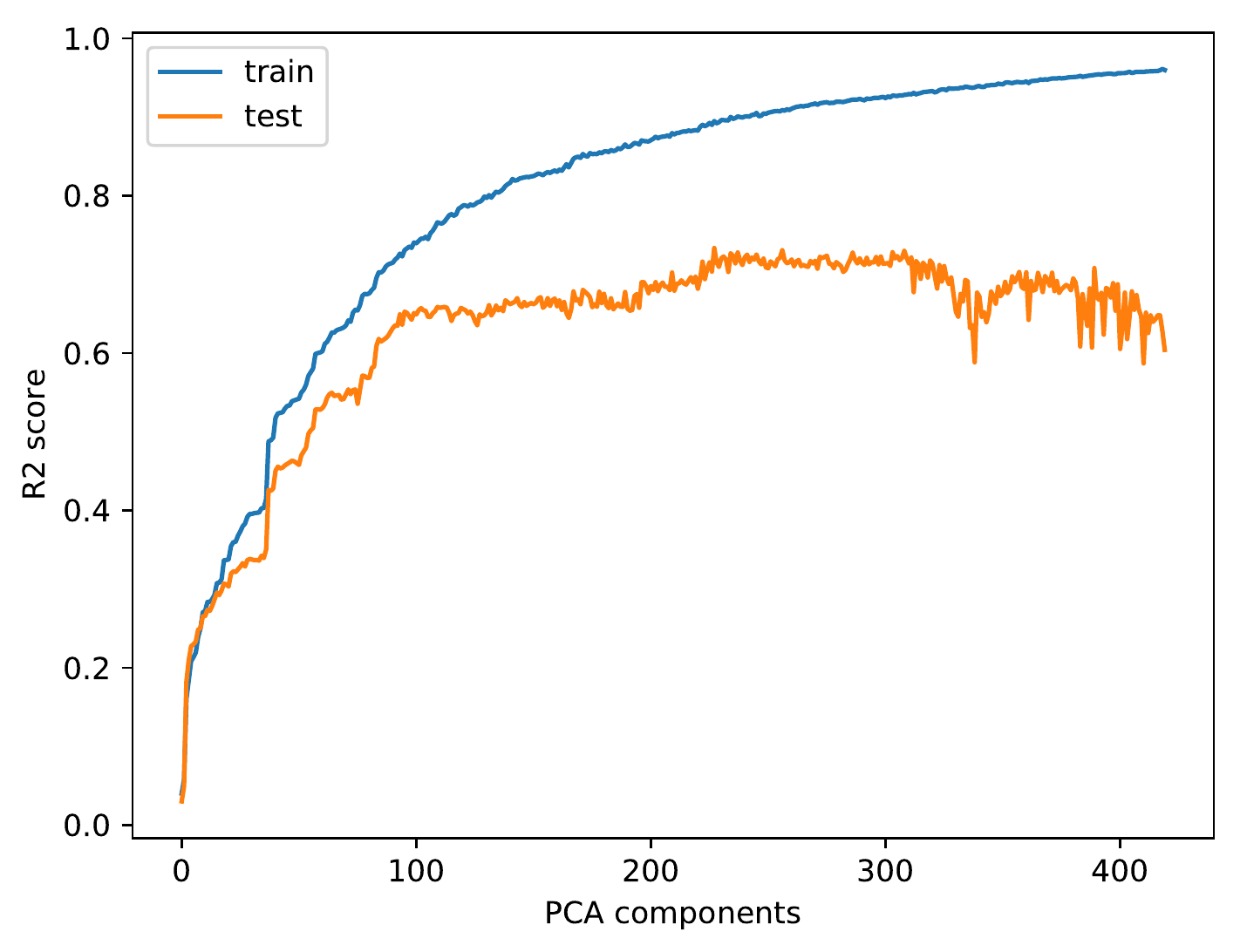}
    \caption{$R^2$ score on Position Data}
    \label{fig:f1}
  \end{subfigure}
  \hfill
  \begin{subfigure}[b]{0.49\linewidth}
    \includegraphics[width=\linewidth]{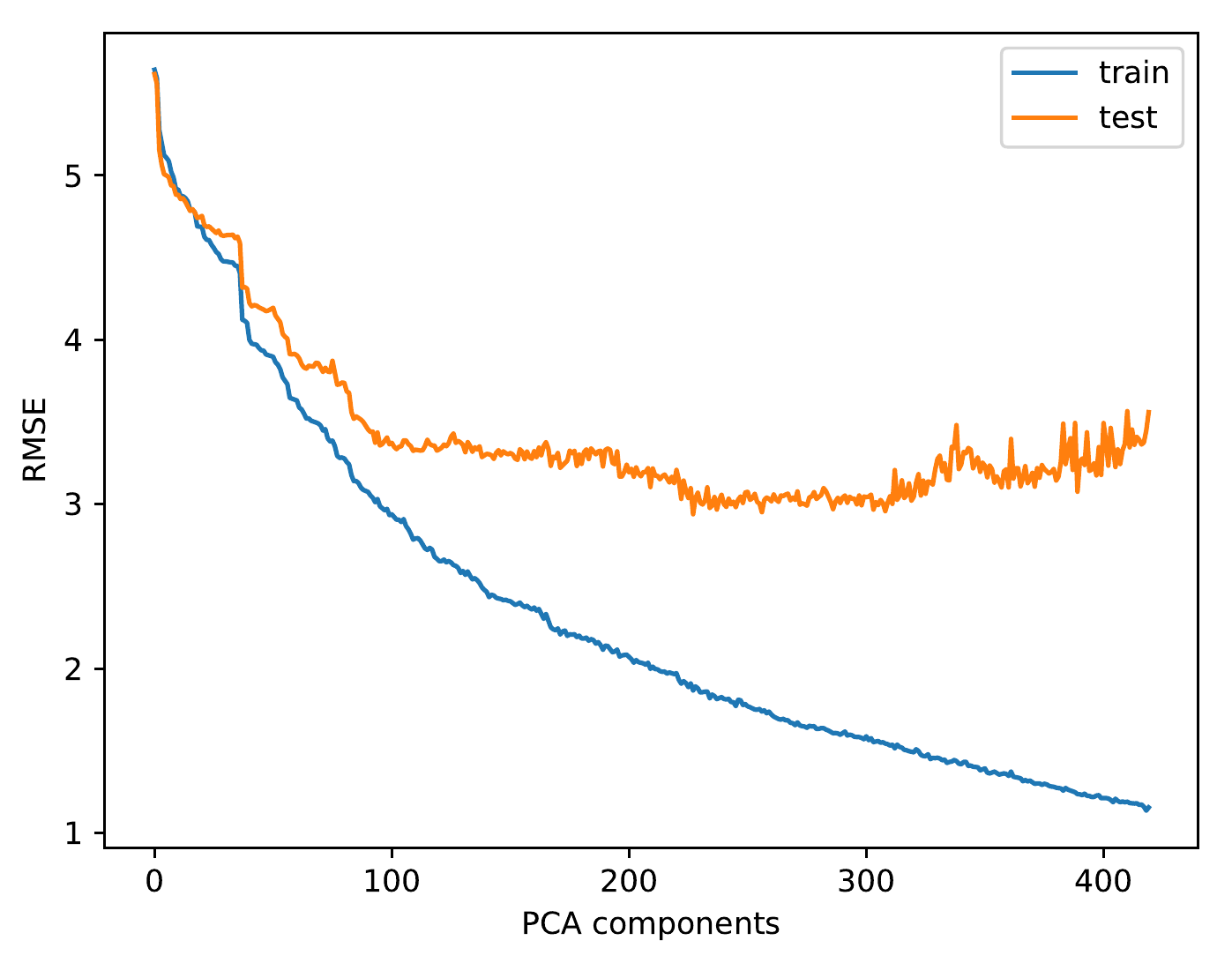}
    \caption{RMSE on Position Data}
    \label{fig:f2}
  \end{subfigure}
  \begin{subfigure}[b]{0.49\linewidth}
    \includegraphics[width=\linewidth]{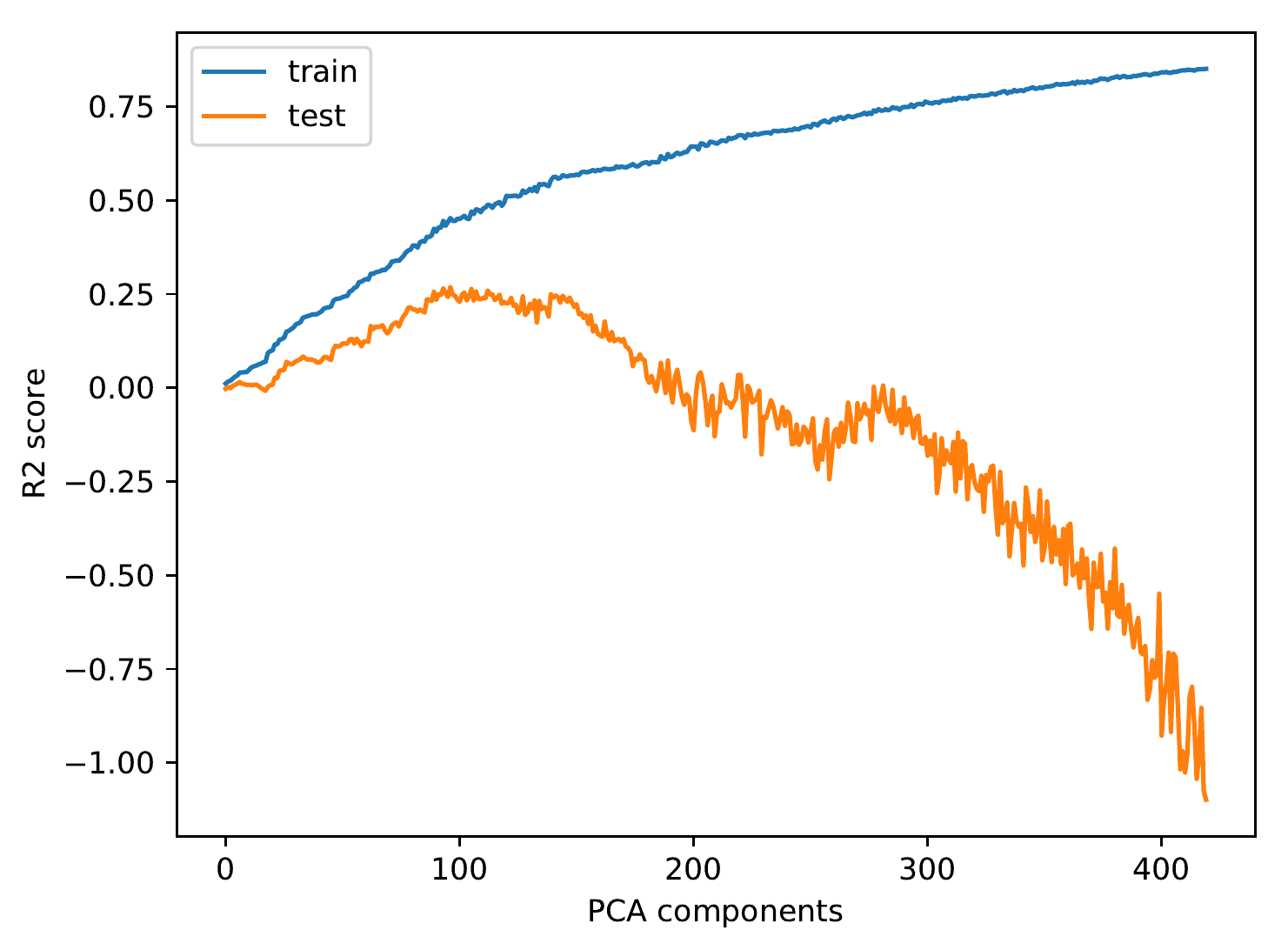}
    \caption{$R^2$ score on Velocity Data}
    \label{fig:f3}
  \end{subfigure}
  \hfill
  \begin{subfigure}[b]{0.49\linewidth}
    \includegraphics[width=\linewidth]{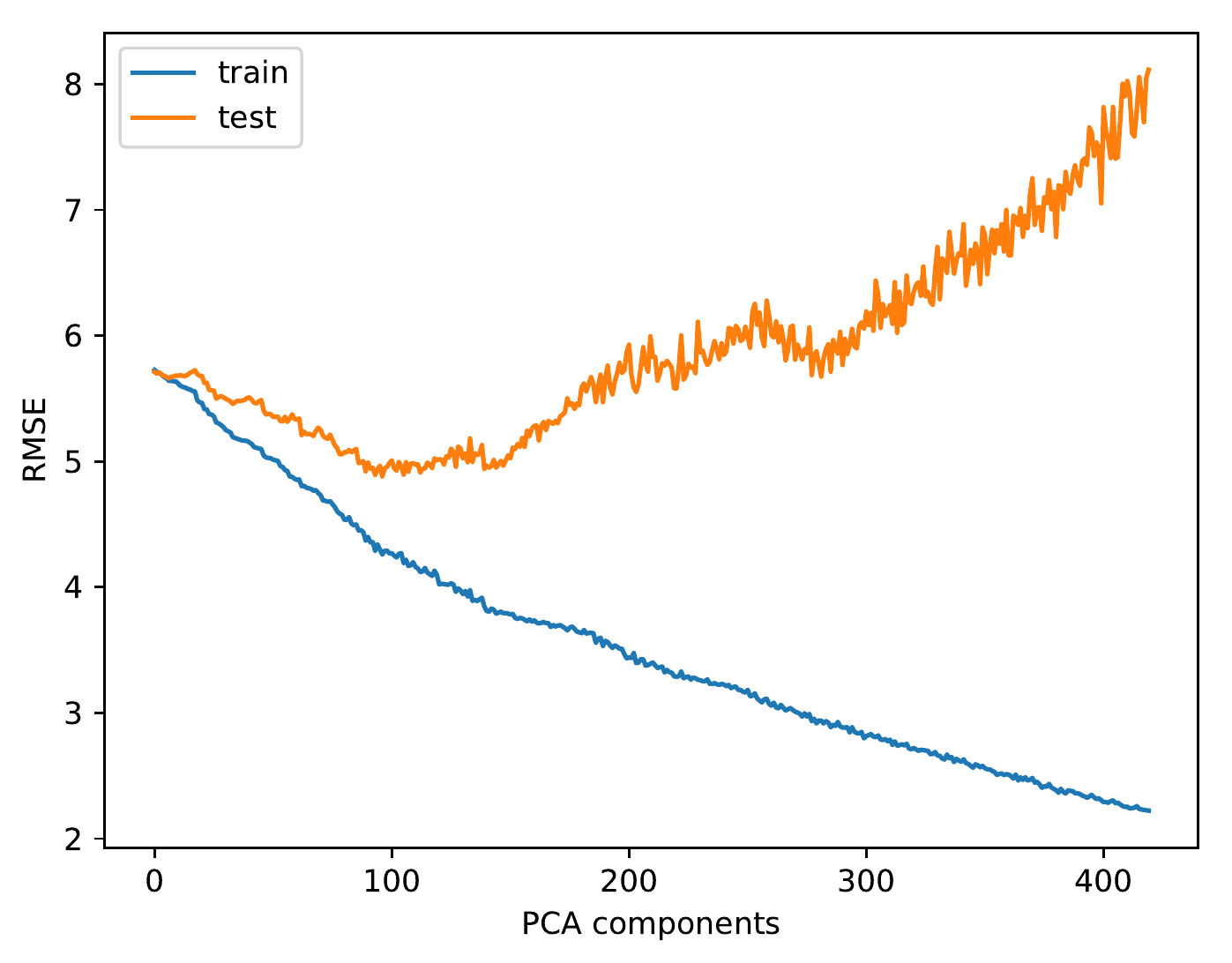}
    \caption{RMSE on Velocity Data}
    \label{fig:f4}
  \end{subfigure}
  
  \caption{PCR Results on EQ}
\end{figure}

From Figs. \ref{fig:f1} and \ref{fig:f2} we can say that using position data for feature extraction (Sec. 3.3), the model started to overfit after having more than around 240 principal components since the RMSE started increasing and $R^2$ score started decreasing for testing data. Similarly, from Figure \ref{fig:f3} and \ref{fig:f4}, we can say that using velocity data for feature extraction, the model started to overfit after having more than around 140 principal components.

% \begin{figure}[!h]
% \subfigure{\includegraphics[width=0.45\linewidth]{}}
% \subfigure{\includegraphics[width=0.45\linewidth]{}}
% \end{figure}

\begin{figure}[h!]
  \begin{subfigure}[b]{0.49\linewidth}
    \includegraphics[width=\linewidth]{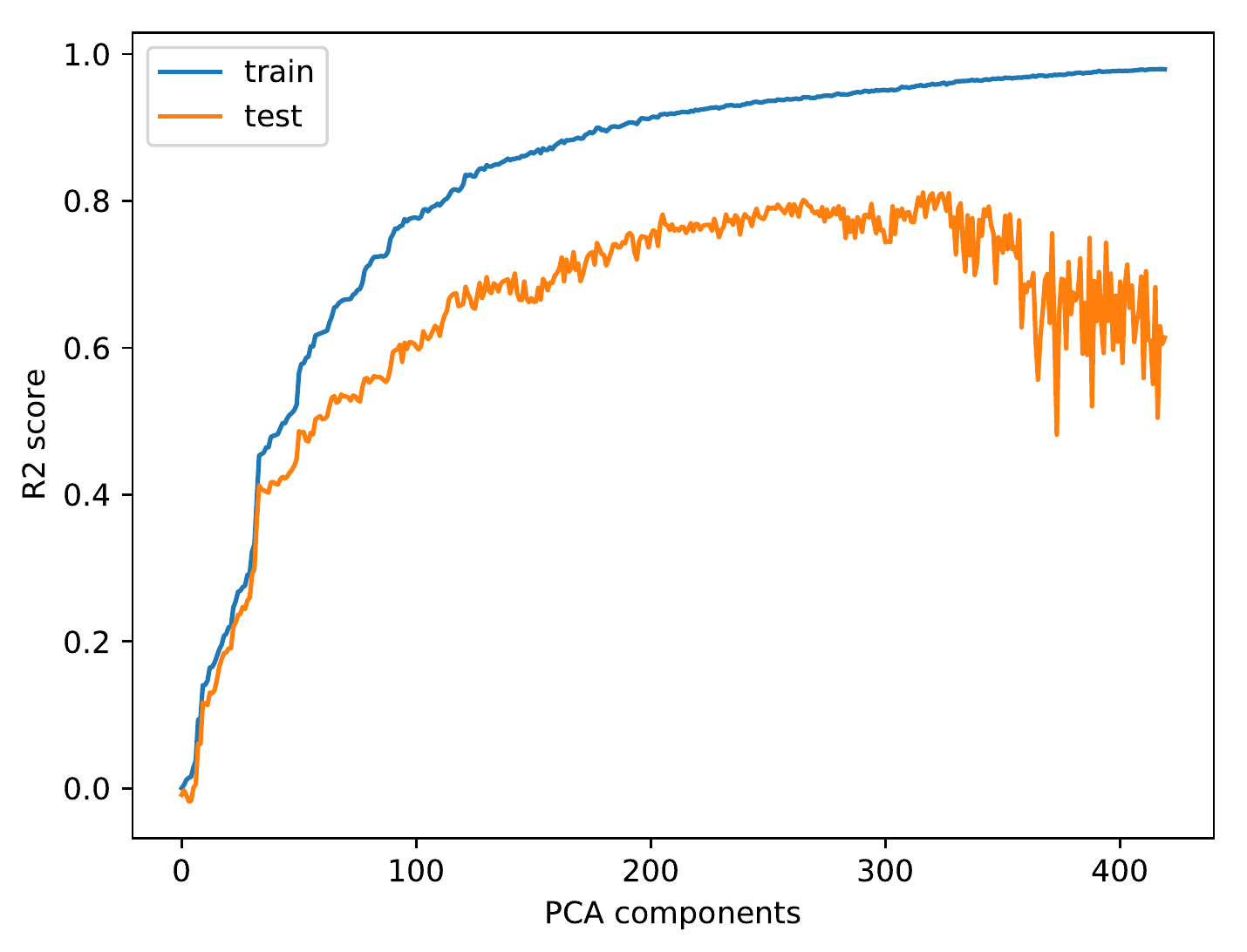}
    \caption{$R^2$ score on Position Data}
    \label{fig:sq_f1}
  \end{subfigure}
  \hfill
  \begin{subfigure}[b]{0.49\linewidth}
    \includegraphics[width=\linewidth]{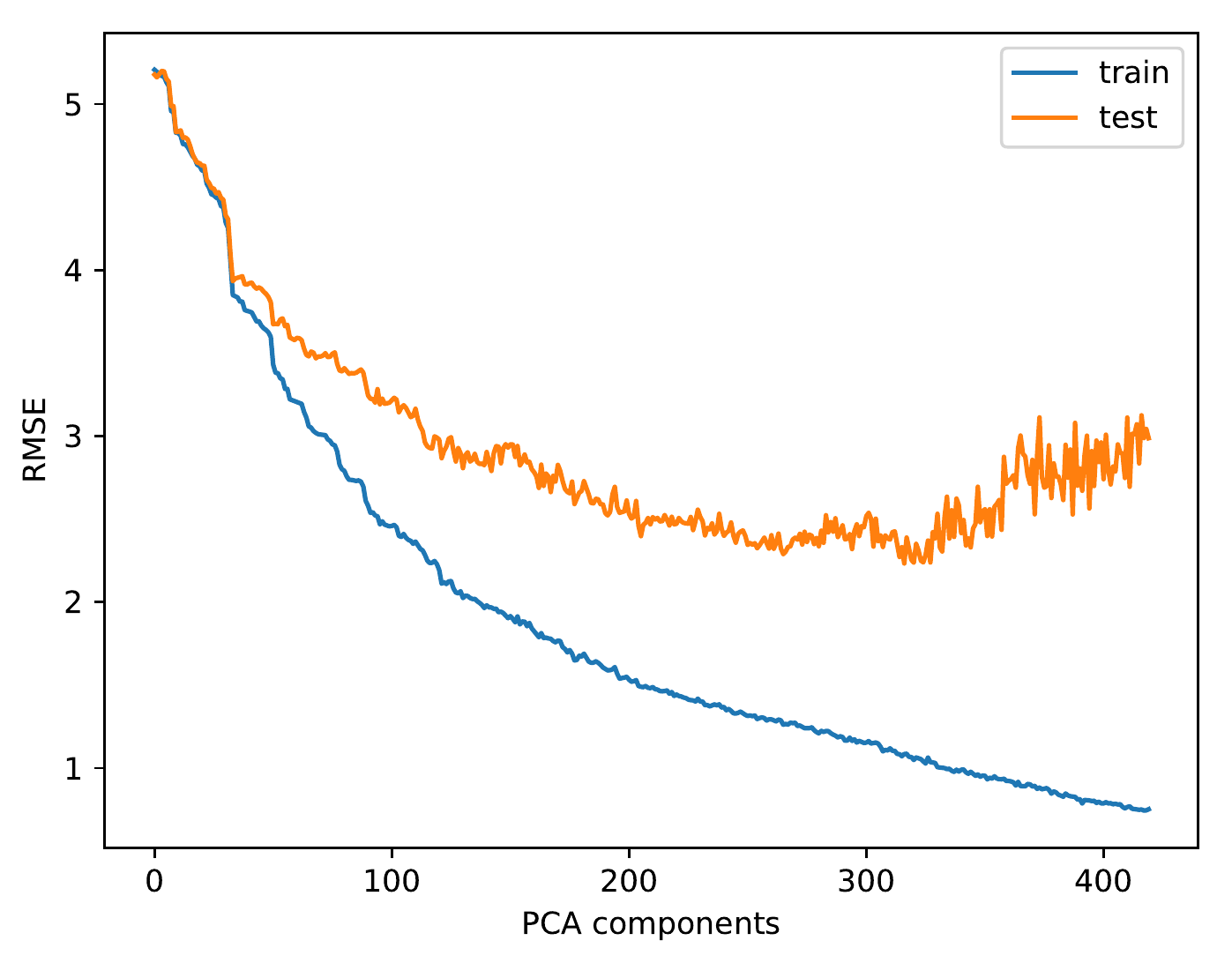}
    \caption{RMSE on Position Data}
    \label{fig:sq_f2}
  \end{subfigure}
  \begin{subfigure}[b]{0.49\linewidth}
    \includegraphics[width=\linewidth]{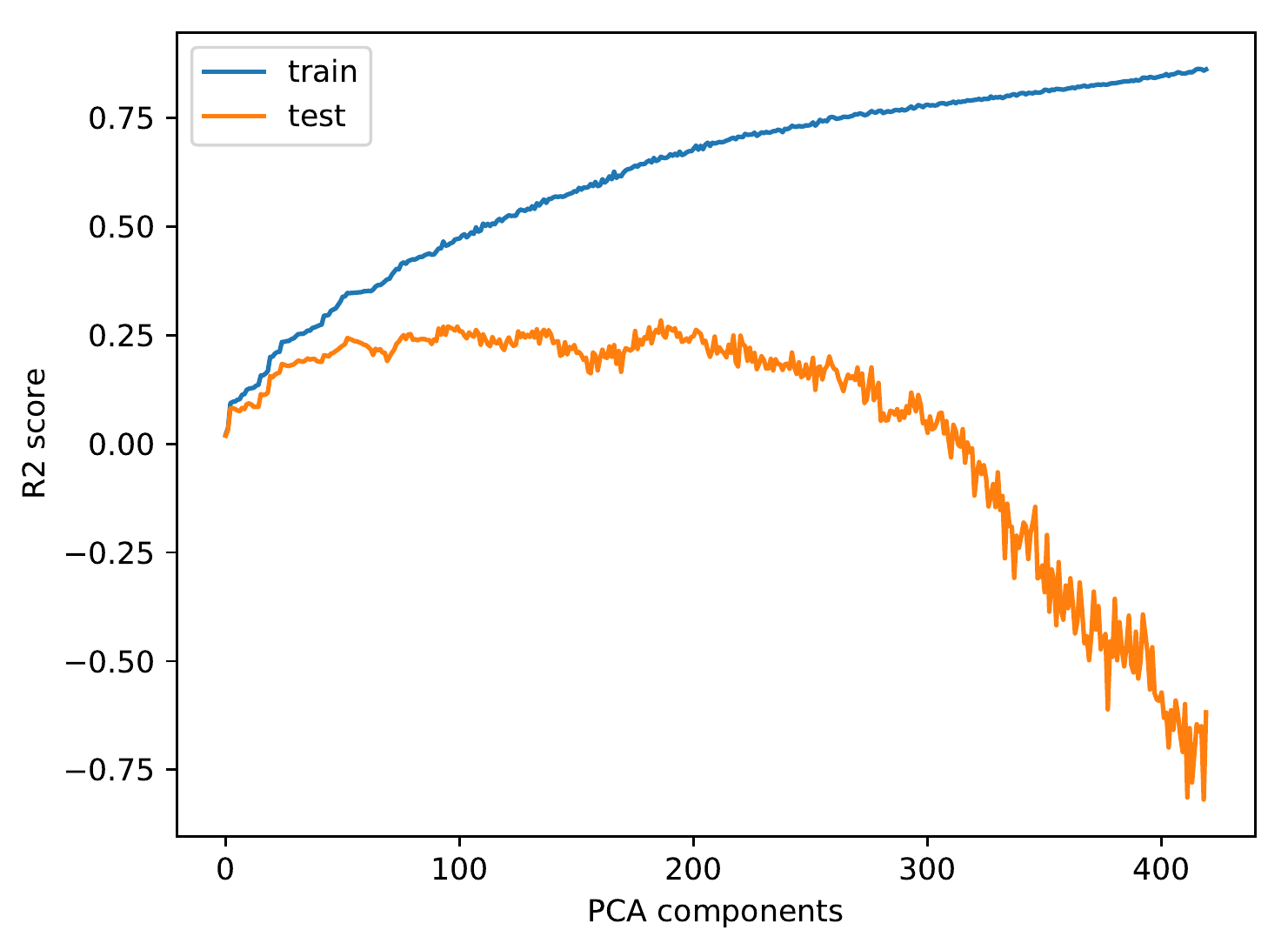}
    \caption{$R^2$ score on Velocity Data}
    \label{fig:sq_f3}
  \end{subfigure}
  \hfill
  \begin{subfigure}[b]{0.49\linewidth}
    \includegraphics[width=\linewidth]{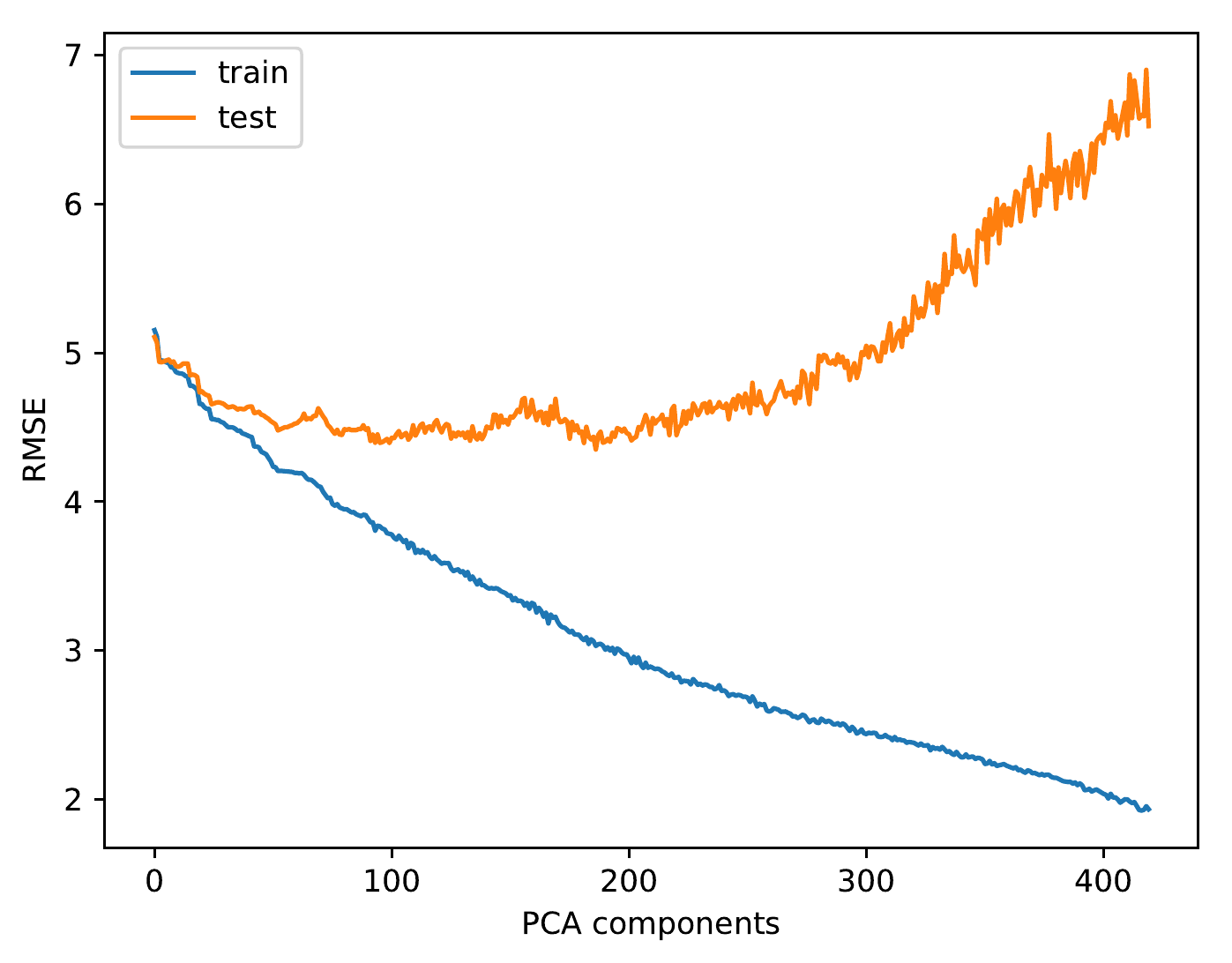}
    \caption{RMSE on Velocity Data}
    \label{fig:sq_f4}
  \end{subfigure}
  
  \caption{PCR Results on SQ}
\end{figure}

\subsubsection*{SQ}
From Figs. \ref{fig:sq_f1} and \ref{fig:sq_f2}, we can say that using position data for feature extraction, the model started to overfit after having more than 260 principal components. Similarly from Figure \ref{fig:sq_f3} \& \ref{fig:sq_f4} when we use velocity data for feature extraction, the model started to overfit after taking more than 170 principal components.
\clearpage
\subsection*{Bayesian Regression}
The model for the Bayesian Regression with the response sampled from a normal distribution is
\begin{equation}
    y \sim N(\beta^Tx, \sigma^2I)
\end{equation}{}
where $\beta$ is the weight vector, $x$ is the feature vector and $\sigma$ is the standard deviation. Here $\beta$ and $\sigma$ are the model parameters. The goal is not to find the single best value of model parameters but to determine the posterior distribution for the model parameters. The posterior probability of the model parameters can be defined as
\begin{equation}
    p(\beta|D) \propto p(D|\beta)p(\beta)
\end{equation}
\begin{equation}
    \beta \sim N(0, \sigma_{\beta}^2I_{d})    
\end{equation}

where  $p(\beta)$ is the initial probability distribution, also known as prior distribution and $p(D|\beta)$ is known as the likelihood function. Using these approaches, we attempted two tasks 1. EQ and SQ Prediction 2. Personality Prediction.

Due to its robustness, it is evident that the model did not overfit on the dataset as the $R^2$ score kept on increasing with the number of principal components. The $R^2$ scores for training and testing set are 0.92 and 0.86 when all the features are considered. We also repeated this experiment by varying the number of principal components incrementally to test its robustness.
\subsubsection*{EQ}
\begin{figure}[h!]
  \begin{subfigure}[b]{0.49\linewidth}
    \includegraphics[width=\linewidth]{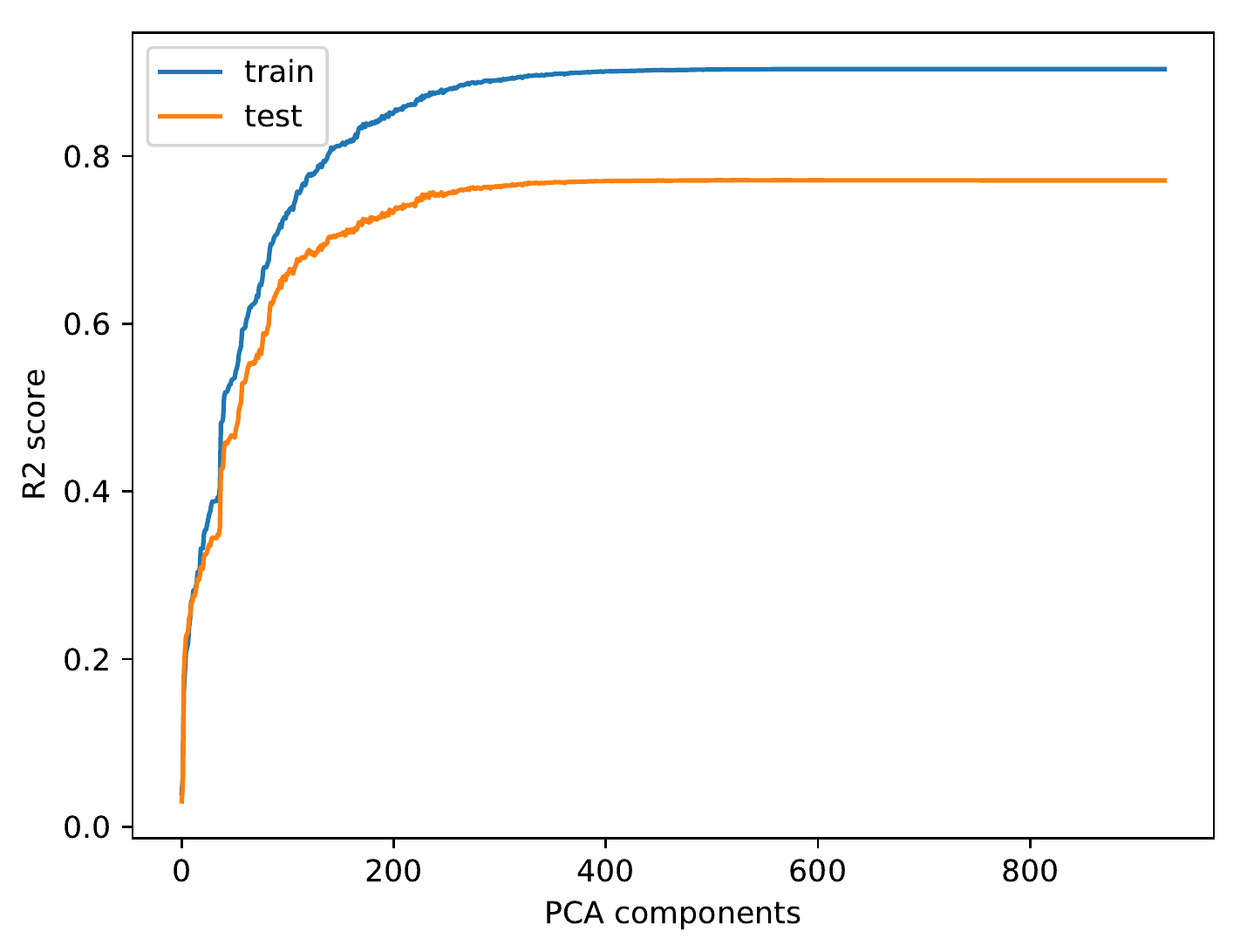}
    \caption{$R^2$ score on Position Data}
    \label{fig:br_f1}
  \end{subfigure}
  \hfill
  \begin{subfigure}[b]{0.49\linewidth}
    \includegraphics[width=\linewidth]{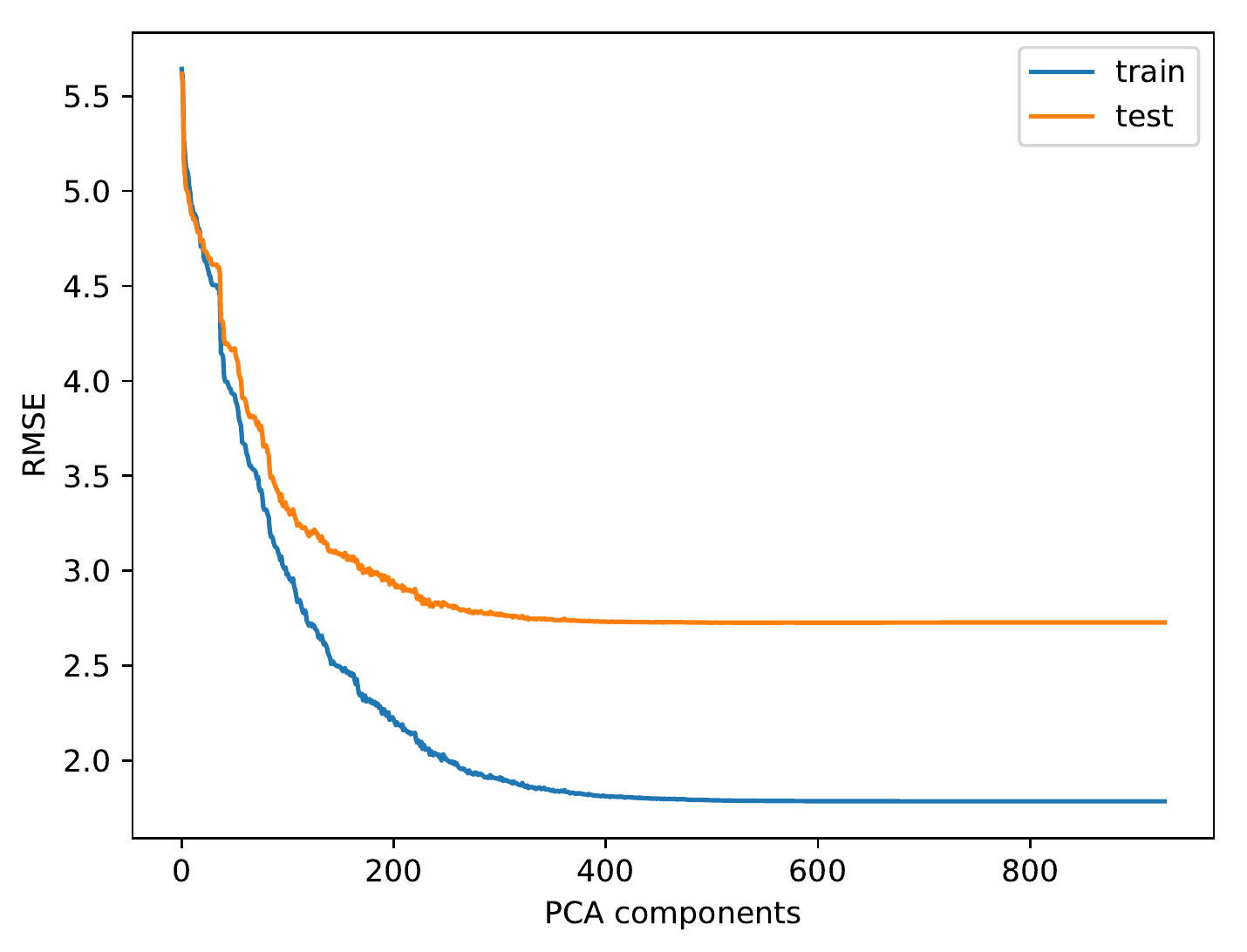}
    \caption{RMSE on Position Data}
    \label{fig:br_f2}
  \end{subfigure}
  \begin{subfigure}[b]{0.49\linewidth}
    \includegraphics[width=\linewidth]{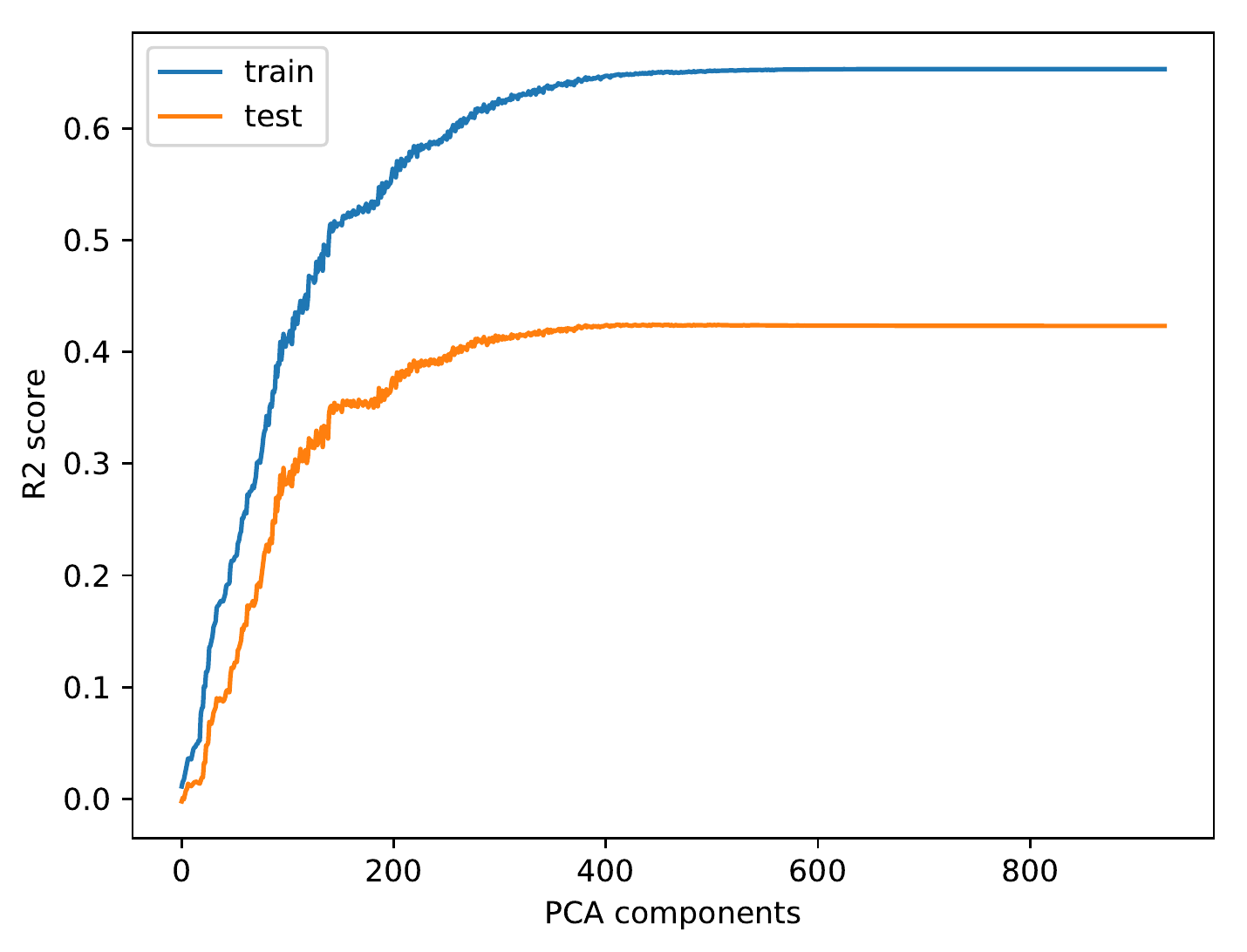}
    \caption{$R^2$ score on Velocity Data}
    \label{fig:br_f3}
  \end{subfigure}
  \hfill
  \begin{subfigure}[b]{0.49\linewidth}
    \includegraphics[width=\linewidth]{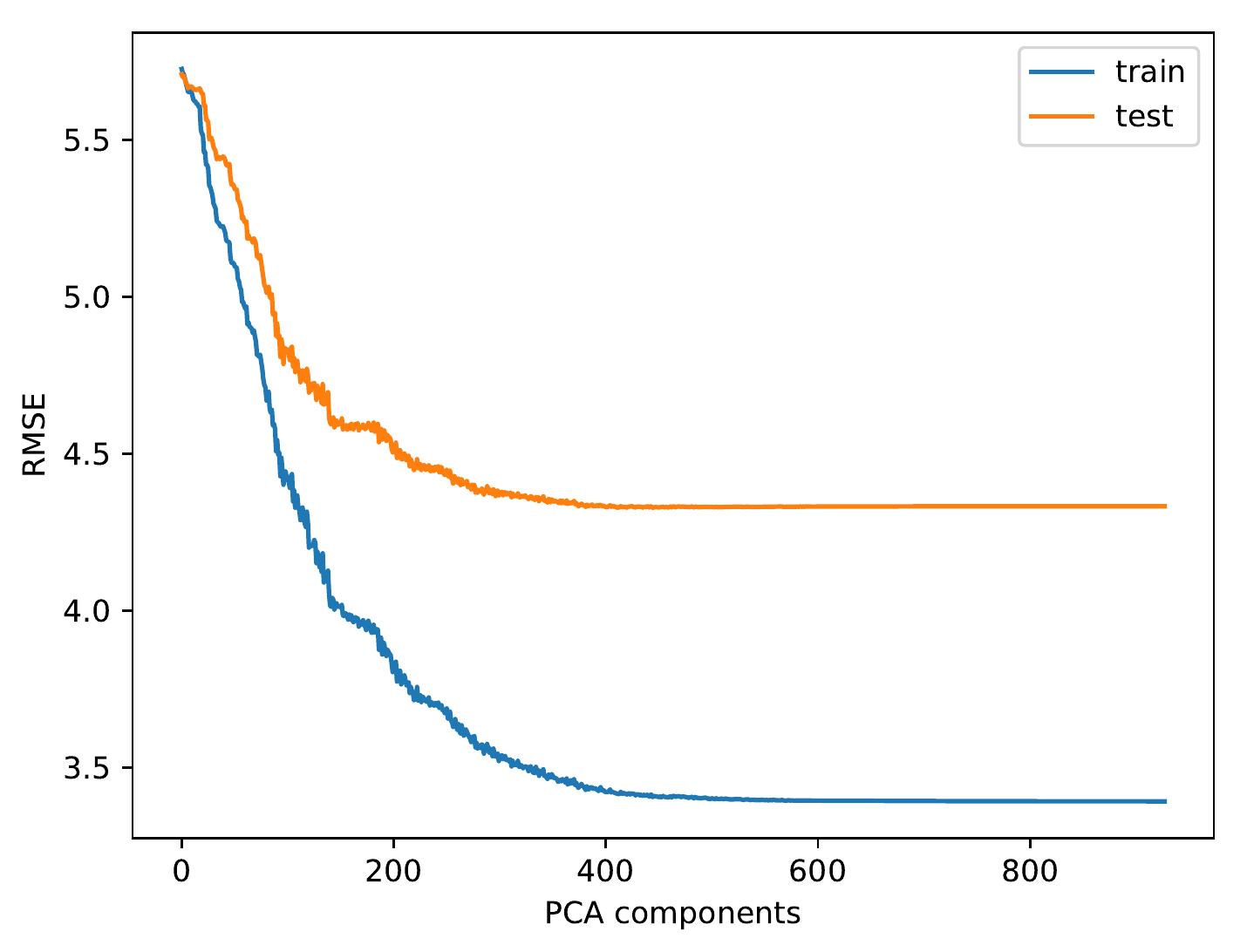}
    \caption{RMSE on Velocity Data}
    \label{fig:br_f4}
  \end{subfigure}
  
  \caption{Bayesian Regression Results on EQ}
\end{figure}

From Figs. \ref{fig:br_f1} \& \ref{fig:br_f2}, we can say that using position data, the maximum $R^2$ score achieved is 0.76 and minimum RMSE is 2.73. From Figs. \ref{fig:br_f3} \& \ref{fig:br_f4}, we can say that  using velocity data, the maximum $R^2$ score is 0.42 and minimum RMSE is 4.35. The RMSE and $R^2$ scores become somewhat saturated at some point but still gets better marginally with the increase in principal components.
\subsubsection*{SQ}
\begin{figure}[h!]
  \begin{subfigure}[b]{0.49\linewidth}
    \includegraphics[width=\linewidth]{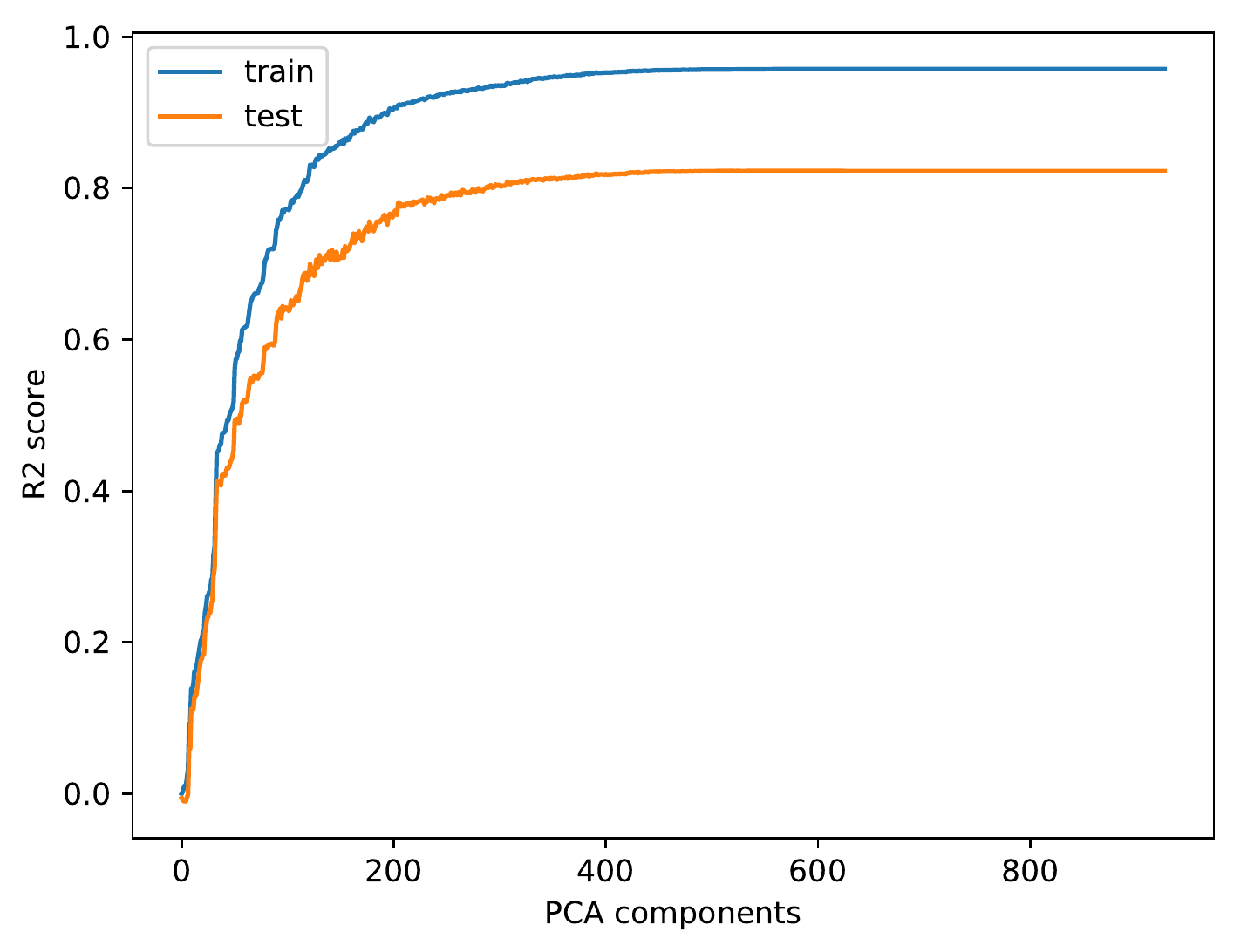}
    \caption{$R^2$ score on Position Data}
    \label{fig:br_sq_f1}
  \end{subfigure}
  \hfill
  \begin{subfigure}[b]{0.49\linewidth}
    \includegraphics[width=\linewidth]{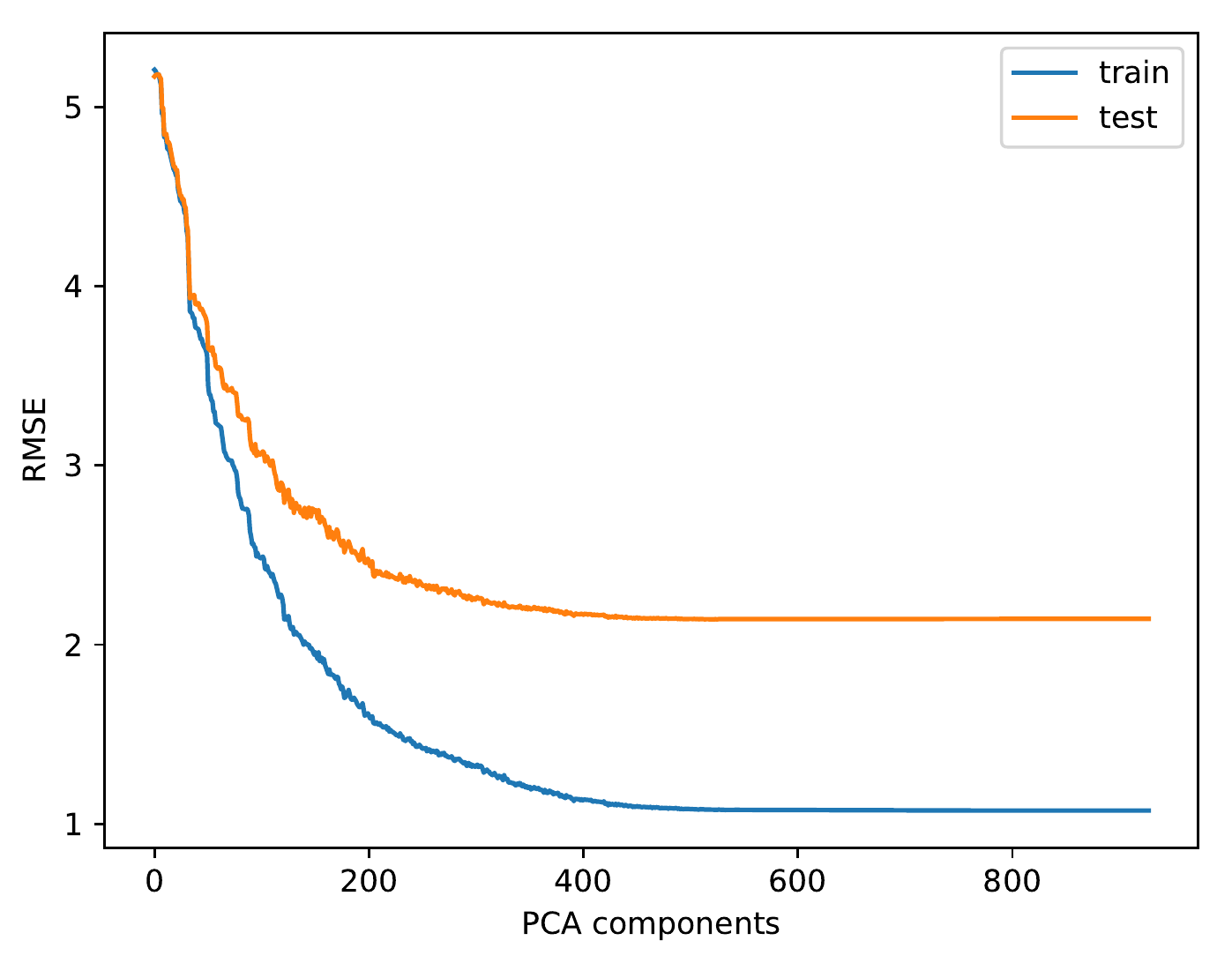}
    \caption{RMSE on Position Data}
    \label{fig:br_sq_f2}
  \end{subfigure}
  \begin{subfigure}[b]{0.49\linewidth}
    \includegraphics[width=\linewidth]{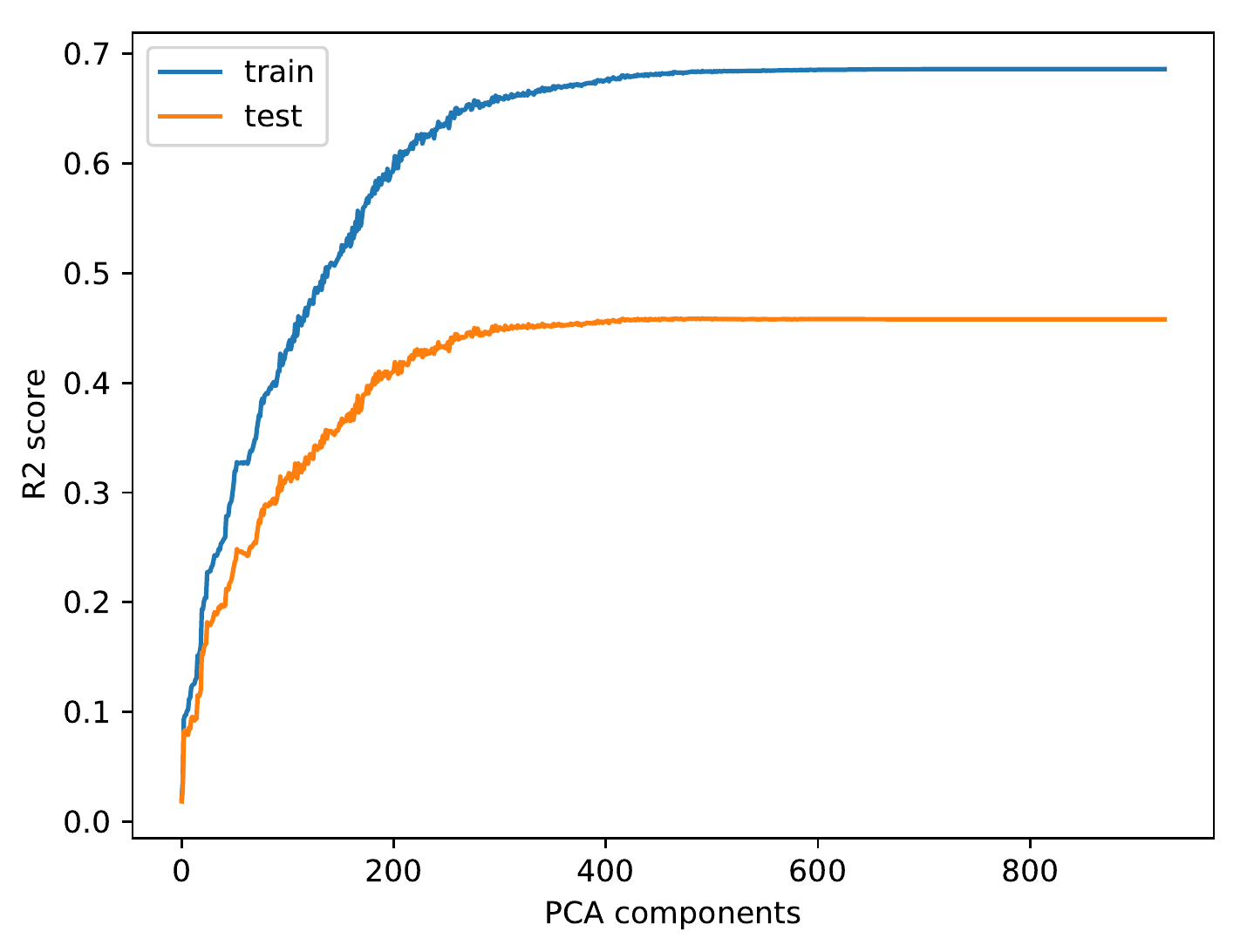}
    \caption{$R^2$ score on Velocity Data}
    \label{fig:br_sq_f3}
  \end{subfigure}
  \hfill
  \begin{subfigure}[b]{0.49\linewidth}
    \includegraphics[width=\linewidth]{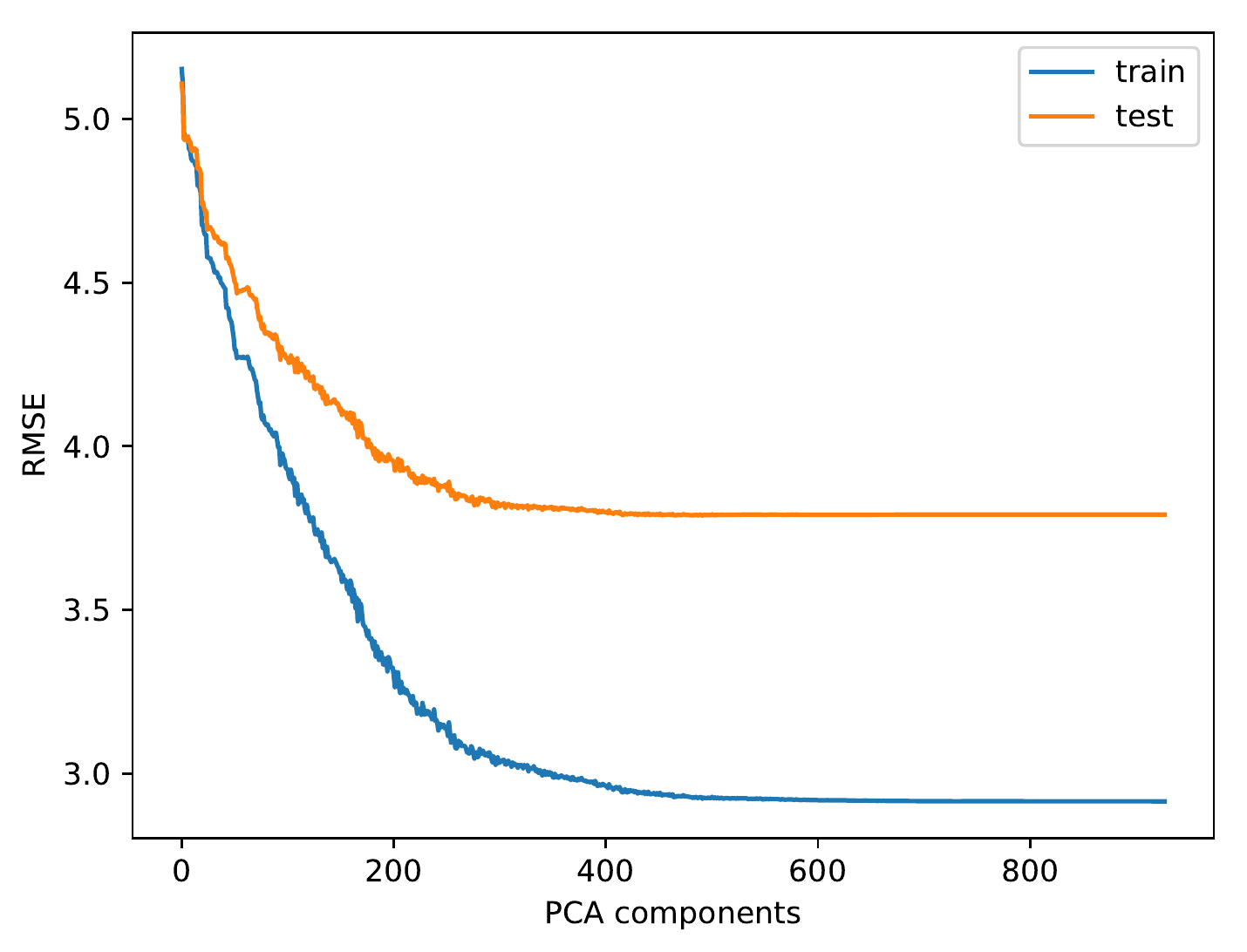}
    \caption{RMSE on Velocity Data}
    \label{fig:br_sq_f4}
  \end{subfigure}
  
  \caption{Bayesian Regression Results on SQ}
\end{figure}

From Figs. \ref{fig:br_sq_f1} \& \ref{fig:br_sq_f2}, we can say that using position data, the maximum $R^2$ score is 0.82 and minimum RMSE is 2.18. From Figs. \ref{fig:br_sq_f3} \& \ref{fig:br_sq_f4}, we can say that using velocity data gained maximum $R^2$ score of 0.44 and minimum RMSE of 3.82. 
% after taking more than 350 principal components, the $R^2$ scores for both training and testing set became saturated. 
In the paper, we reported the RMSE and $R^2$ scores of Bayesian Regression without taking any principal components. 

In EQ, using position data, the RMSE and $R^2$ is 2.72 and 0.77. Similarly, using velocity data, the RMSE and $R^2$ is 4.43 and 0.42. In SQ, using position data, the RMSE and $R^2$ is 2.16 and 0.86. Similarly, using velocity data, the RMSE and $R^2$ is 3.83 and 0.46. Since, the results were close but better than that of after performing the PCA, we present all the results in our paper for Bayesian Regression without performing PCA.


\subsection*{Correlation between BFI Personality Traits}

\begin{table}[h!]
\setlength{\tabcolsep}{12pt}
\begin{tabular}{|c|cccc|}
% \begin{tabular}{|p{1cm}|p{1cm} p{1cm} p{1cm} p{1cm}|}
\hline
%  & Openness & Conscientiousness & Extraversion & Agreeableness \\ \hline
  & O & C & E & A \\ \hline
C & -0.093 & - & - & - \\
E & -0.003 & 0.128 & - & - \\
A & 0.021 & \cellcolor[HTML]{C0C0C0}0.341 & \cellcolor[HTML]{C0C0C0}0.358 & - \\
N & 0.217 & \cellcolor[HTML]{EFEFEF}-0.289 & -0.225 & \cellcolor[HTML]{EFEFEF}-0.292 \\ \hline
\end{tabular}
\begin{tabular}{
>{\columncolor[HTML]{EFEFEF}}l 
>{\columncolor[HTML]{C0C0C0}}l }
p\textless{}0.05 & p\textless{}0.01
\end{tabular}
\caption{Results of the Spearman Correlation between the BFI personality traits.}
\label{table:ocean_corr1}
\end{table}